\documentclass[sigconf]{acmart}

\usepackage{algorithm}
\AtBeginDocument{%
  }

\copyrightyear{2025}
\acmYear{2025}
\setcopyright{cc}
\setcctype{by}
\acmConference[SCF '25]{ACM Symposium on Computational Fabrication}{November 20--21, 2025}{Cambridge, MA, USA}
\acmBooktitle{ACM Symposium on Computational Fabrication (SCF '25), November 20--21, 2025, Cambridge, MA, USA}
\acmDOI{10.1145/3745778.3766665}
\acmISBN{979-8-4007-2034-5/2025/11}

\begin{document}

\title[Hierarchical Discrete Lattice Assembly]{Hierarchical Discrete Lattice Assembly: An Approach for the Digital Fabrication of Scalable Macroscale Structures}

\author{Miana Smith}
\authornote{Both authors contributed equally to this research.}
\orcid{REPLACE}
\affiliation{%
\institution{Center for Bits and Atoms}
  \institution{Massachusetts Institute of Technology}
  \city{Cambridge}
  \state{Massachusetts}
  \country{USA}}
\email{miana@mit.edu}

\author{Paul Arthur Richard}
\authornotemark[1]
\affiliation{%
  \institution{École Polytechnique Fédérale de Lausanne}
  \city{Lausanne}
  \country{Switzerland}
}
\email{paul.richard@epfl.ch}

\author{Alexander Htet Kyaw}
\affiliation{%
\institution{School of Architecture and Planning}
  \institution{Massachusetts Institute of Technology}
  \city{Cambridge}
  \state{Massachusetts}
  \country{USA}}
\email{alexkyaw@mit.edu}

\author{Neil Gershenfeld}
\affiliation{%
\institution{Center for Bits and Atoms}
  \institution{Massachusetts Institute of Technology}
  \city{Cambridge}
  \state{Massachusetts}
  \country{USA}}
\email{gersh@cba.mit.edu}

\renewcommand{\shortauthors}{Smith et al.}

\begin{abstract}
  Although digital fabrication processes at the desktop scale have become proficient and prolific, systems aimed at producing larger-scale structures are still typically complex, expensive, and unreliable. In this work, we present an approach for the fabrication of scalable macroscale structures using simple robots and interlocking lattice building blocks. A target structure is first voxelized so that it can be populated with an architected lattice. These voxels are then grouped into larger interconnected blocks, which are produced using standard digital fabrication processes, leveraging their capability to produce highly complex geometries at a small scale. These blocks, on the size scale of tens of centimeters, are then fed to mobile relative robots that are able to traverse over the structure and place new blocks to form structures on the meter scale. To facilitate the assembly of large structures, we introduce a live digital twin simulation tool for controlling and coordinating assembly robots that enables both global planning for a target structure and live user design, interaction, or intervention. To improve assembly throughput, we introduce a new modular assembly robot, designed for hierarchical voxel handling. We validate this system by demonstrating the voxelization, hierarchical blocking, path planning, and robotic fabrication of a set of meter-scale objects.  
\end{abstract}




\begin{teaserfigure}
  \includegraphics[width=\textwidth]{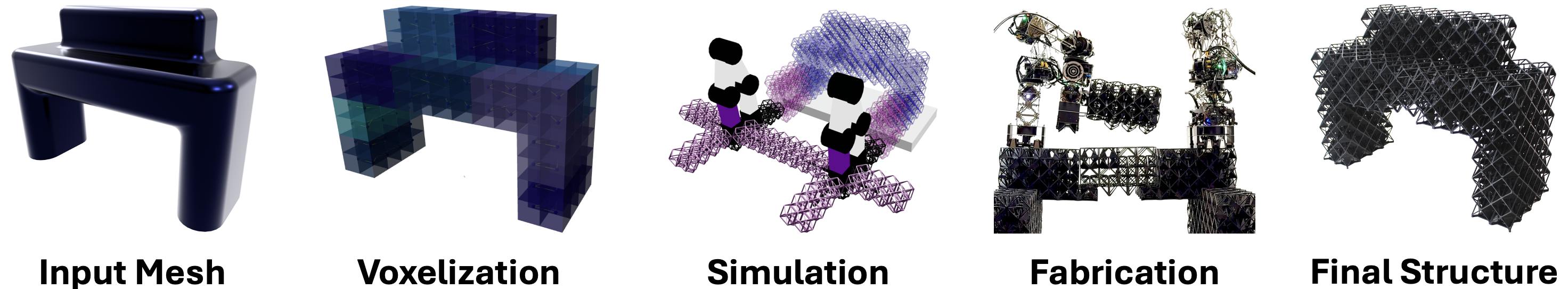}
  \caption{Pipeline from STL Mesh to voxelization, to simulation, to robotic assembly, to the final structure for a small bench.}
  \label{fig:teaser}
\end{teaserfigure}


\maketitle

%% MAIN BODY 

\section{Introduction}

Machines for digital fabrication typically cannot make objects or structures larger than themselves. Although this is not necessarily limiting on the desktop scale, as target structures scale toward human or architectural scales, the prospect of ever-growing machine footprints eventually becomes untenable \cite{meisel_design_2022}. Current digital fabrication approaches for larger-scale structures often section the target structure into parts that can be produced separately and then assembled, usually manually \cite{formlabs2023howto} \cite{kovacs_trussfab_2017}, \cite{baudisch_kyub_2019}. Approaches that consider assembly automation often rely on large and expensive industrial robotic arms with limited overall mobility \cite{menges_fabricate_2017}. In either case, there is no clear strategy toward scaling to unlimited build footprints or autonomous environments. 

In contrast, collective robotic construction, in which a group of robots collaborate to assemble a structure larger than any individual robot, offers significant promise towards scale-agnostic, efficient, and autonomous fabrication \cite{petersen_review_2019}. However, hardware demonstration of these systems has typically been limited in scale \cite{jenett_materialrobot_2019}, load capacity \cite{petersen2011termes}, or geometric freedom \cite{melenbrink_autonomous_2021}. This is because automated large-scale fabrication with mobile robots is inherently challenging, with many competing priorities, such as structural stability versus mass efficiency, geometric complexity versus ease of robotic fabrication, or robotic functionality versus swarm simplicity.  

By designing a material system together with its robotic assembly system, we can address some of these issues. The mechanical interplay between the assembly robots and the underlying material system can enable local error correction, allowing relatively simple robots to build large and precise structures \cite{jenett_materialrobot_2019}. In this work, we present an approach to the assembly of large structures using small relative robots that manipulate hierarchical blocks of lattice material. Our approach uses a new type of discrete lattice block and robotic system aimed at improving assembly throughput and mechanical stability. We use standard digital fabrication processes to pre-produce compounded blocks of architected lattices– a high efficiency material system– at the size scale of tens of centimeters, and then use mobile robots to take over beyond that, to the meter scale. To ensure structural stability, we developed a tool similar to a 3D printing slicer that imports a 3D mesh, voxelizes it, and finds interleaved patterns for robots to assemble. It then simulates the assembly live, streaming instructions to the hardware system. Our contributions are: 
\begin{itemize}
  \item An interactive tool for voxelizing 3D shapes and finding connected patterns for different shapes of blocks.
  \item A digital twin path planning and simulation environment integrated with the hardware system, for both assembly control and live design and re-design of structures. 
  \item The development of a compounded set of self-aligning, interlocking, and load bearing octet lattice blocks designed for robotic assembly.
  \item The development of a new modular inchworm style assembly robot, designed for hierarchical assembly and low system cost and complexity. 
\end{itemize}

\section{Related Work}

Our approach builds on existing research in scalable building systems, robotic construction and assembly systems, and material-robot and modular robotic systems. In this section, we overview these fields and identify trends and limitations that guided our approach in this work.  

\subsection{Scalable Building Systems}

Although this work is focused on the robotic assembly of structures, scalable prototyping systems designed for human use offer useful insights into reliable mechanical design for building at larger scales. Specifically, we draw from systems that are not significantly limited by the build volume of the fabrication systems used. Broadly, these systems tend to fall into three categories: continuous, sectioned, and discrete fabrication processes. 

\subsubsection{Continuously Fabricated Systems}

An example of a continuous fabrication system capable of producing meter-scale structures is Protopiper \cite{agrawal_protopiper_2015}, which extrudes tape-based pipes to draw out human- to architecture- scale structures. However, this system has limited load bearing capacity. Wire benders, such as \cite{pensalabs2023}, are similarly able to extrude much larger structures than the machine itself, and with potentially good load-bearing capacity, such as in \cite{BHUNDIYA-bendforming}. However, these continuous extrusion based processes have no mechanisms for preventing error accumulation in the built structure, and instead rely on the precision of the machine, or human intervention, to correct problems as they arise— a potentially significant issue, given that a small error in bend angle over a long distance can result in substantial deflection from the target shape.

\subsubsection{Sectioned Fabrication and Assembly}
Splitting up a target structure into subcomponents can help ameliorate some of the issues with error accumulation, while additionally opening up a wider array of fabrication systems. These approaches are typically geared toward breaking a larger model down into either 2D components that can be cut and assembled together, or smaller 3D models that can be separately fabricated and assembled. LuBan3D \cite{luban3d2025} is a software package aimed at doing exactly this: facilitating users to fabricate structures much bigger than the machines they have, such as the house in \cite{sass2006instant}. More specialized examples aimed at non-expert users include Kyub \cite{baudisch_kyub_2019}, which enables the design and automatic creation of load-bearing structures made from laser cut plywood, or HingCore \cite{10.1145/3526113.3545618} and PopCore \cite{10.1145/3639473.3665787}, which offer joinery-free methods for processing foamcore or equivalent sandwich panel material. The primary issue with these methods is that they produce highly specific parts for specific designs, often requiring dexterous methods for assembly. This makes automation much harder: handling extremely diverse geometries in many degrees of freedom is a hard problem in robotics. Additionally, this typically prevents re-use of parts, as components are likely not portable to new designs, requiring full re-fabrication and re-assembly to make design iterations. 

\subsubsection{Discrete Assembly}
Continuing the trend toward breaking up a big structure into smaller parts is the class of systems that use discrete assembly, where a single part type (or a small library of component types) is used to iteratively build many different target structures. A classic example of this is LEGO®, which enables an imprecise assembler (e.g. a child) to build precise and relatively robust structures. This is because the material system is designed to self-align and error correct during the assembly process, so that the accuracy of the final structure is primarily determined by the tolerances in the part manufacturing, and not by the tolerances of the assembler, in contrast to continuous extrusion fabrication processes. LEGO® uses a volumetric brick-based decomposition, but other forms exist, including strut-and-node systems \cite{kovacs_trussfab_2017}, facet-based assemblies \cite{jenett_building_2018}, and voxel-based assemblies \cite{gregg_ultra-light_2018}. This style of system lends itself well to robotic fabrication, as it features a limited and similar set of parts, often designed to accommodate imprecise assemblers. 

Among discrete assembly approaches for large-scale structures, we focus on using discrete architected lattices, as these systems offer competitive mechanical properties at a lightweight \cite{schaedler_architected_2016}. This is desirable both to improve the performance of the resultant structure and to lower the payload demands on an assembly robot, which reduces some engineering challenges associated with heavy building materials \cite{goessens_feasibility_2018}. Discrete versions of architected lattices have demonstrated extreme performance results \cite{cheung_reversibly_2013}, designed material anisotropies \cite{jenett_discretely_2020}, and integrated electronic functionality \cite{smith_voxel_2025}. Manually, these have been assembled into a wide array of meter-scale structures, such as morphing aero- and hydro- structures \cite{jenett_digital_2017} \cite{parra_rubio_modular_2023}, vehicles \cite{jenett_discrete_2020}, or load-bearing static structures \cite{smith_voxel_2025}, demonstrating the broad applicability that is desirable for a generalist building system. 

\subsection{Robotic Fabrication Systems}

We now consider the question of how to automate the building of large structures. Approaches for this, again, roughly fall into three categories: large machines to build large structures, large robots to assemble large structures, or small mobile robots to assemble large structures. 

\subsubsection{Static Gantries}

In the first approach, a typically gantry-based machine is used to assemble a structure,  as in \cite{apolinarska2016sequential}, or, more commonly, to 3D print a structure,  as shown in \cite{batikha_3d_2022} often using systems in the style of \cite{cobod_eu_nodate}, for example. Though these approaches have been demonstrated at commercial architectural scales, they are still fundamentally limited by the need for very large machines, which introduces both an upper limit on the size of the structure that can be made, as well as significant logistics challenges, either in terms of transporting parts, in the case of pre-fabrication, or building the machine in place, in the case of in-situ fabrication. With the extrusion based approaches, there are additional open challenges in material formulation and performance \cite{marchment_mesh_2020} \cite{roux_life_2023}. For these reasons, we are interested in further exploring assembly based approaches. 

\subsubsection{Industrial Arms}

At the large, complex, and expensive side of the assembly based approaches are projects that have focused on using industrial robotic arms for assembly. These have demonstrated a range of material types, such as steel rebar \cite{Ma-metal-frame}, timber \cite{APOLINARSKA2021103569}, or bricks \cite{GHARBIA2020101584}, including commercial systems such as \cite{noauthor_hadrian_nodate} or \cite{noauthor_sam_nodate}. However, these systems have limited accessible footprints, based on the reach of the robotic arm (or the linear axis it is mounted on), or feature significant localization challenges, if on a mobile base \cite{bodea_additive_2022}. These projects often rely on the high performance of their robotic and sensing systems to compensate for the simplicity in their building materials— by shifting some of the complexity onto the material system, we can reduce some of the demands (and size and cost) on the robotic system. 

\subsubsection{Material-Robot Systems}

Building systems that rely on the co-design and interaction of the robots and material systems are often referred to as material-robot systems \cite{jenett_materialrobot_2019}. The degree of material-robot integration varies across the literature. At one extreme end, self-reconfiguring modular robots, as in \cite{neubert_soldercubes_2016} or \cite{HAUSER2020103467}, could be considered a material-robot system where there is no distinction between the building system and the robot, resulting in a very complex material system. At the other end, systems such as \cite{goessens_feasibility_2018} or \cite{leder_leveraging_2022} only shift a minimal amount of complexity onto the material system. 

For discrete lattices, systems for voxel-based assembly, such as in \cite{jenett_materialrobot_2019}, \cite{smith_self-reconfigurable_2024}, or \cite{gregg_ultralight_2024}, as well as strut-based assembly, such as in \cite{hsu_application_2016} or \cite{yoon_shady3d_2007} have been explored. \cite{jenett_materialrobot_2019} demonstrated that an inchworm style robot could accurately traverse and assemble a voxel structure with no global feedback, based on the error correction between the voxels and the robot. However, this system relies on magnets and an underlying steel table to form connections and provide mechanical stability, and has low load bearing capacity on its own. This system is evolved in \cite{park_soll-e_2023} to demonstrate the assembly of 100s of voxels \cite{gregg_ultralight_2024}; however, the complexity of the robotic system is substantially increased, requiring three separate high degree of freedom robotic systems to perform assembly (a voxel carrying robot, a voxel installing robot, and a voxel fastening robot internal to the lattice), while the size scale of the lattice limits shape fidelity (300mm pitch).  

In both of these voxel assembly systems, as the structure grows, the assembly throughput decreases, as the robot must traverse the built structure back and forth to pick up and install new material. \cite{abdel-rahman_self-replicating_2022} proposes that recursive hierarchical systems can help improve this issue— if assembly robots can assemble more assembly robots, then the swarm can increase its own parallelization, and if assembly robots can manipulate larger quantities of material at once, then they can build more efficiently. \cite{smith_self-reconfigurable_2024} demonstrates a first version of the hardware for a load-bearing modular voxel assembler, but is limited to desktop-scale objects. 

\subsubsection{Synthesis}

Based on the prior art, we can distill some guidelines for an ideal robotic assembly-based digital fabrication system. Collective robotic approaches that have demonstrated the most promise towards scaling to practical applications typically strike a careful balance between the complexity of the material system and robotic system. The material system needs to be kept simple enough to be manufactured, while sophisticated enough to account for limited robotic functionality. The robot, in turn, needs enough features and degrees of freedom to reliably traverse and manipulate the lattice, but not so many that it becomes untenable to increase the amount of robots used. An ideal connection system only requires access from a single direction, without adding significant installation time. To maintain effective assembly times for larger structures, as well as to improve reliability, the system, as much as possible, should be able to parallelize its function and offer hierarchical assembly. 

\section{System Overview}

\begin{figure}[h]
  \centering
  \includegraphics[width=\linewidth]{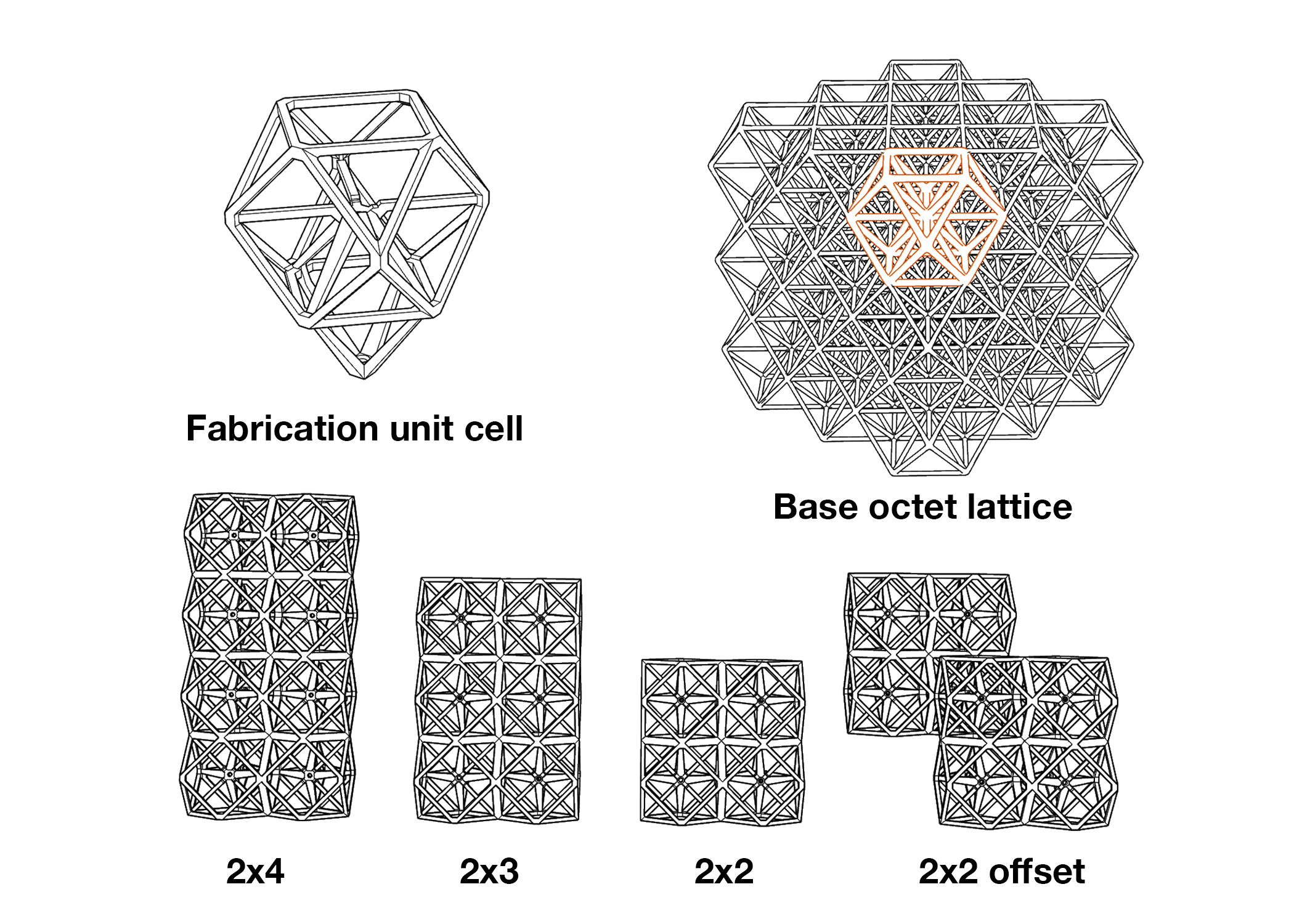}
  \caption{An overview of the lattice type and building blocks used in this project. The basic lattice type is an edge-connected octet lattice, which is decomposed into an extended cuboctahedron-octet, which is then compounded into different arrangements for robotic assembly.}
  \Description{Graphical overview of lattice block types.}
  \label{fig:lattice-types}
\end{figure}

\begin{figure}[h]
  \centering
  \includegraphics[width=\linewidth]{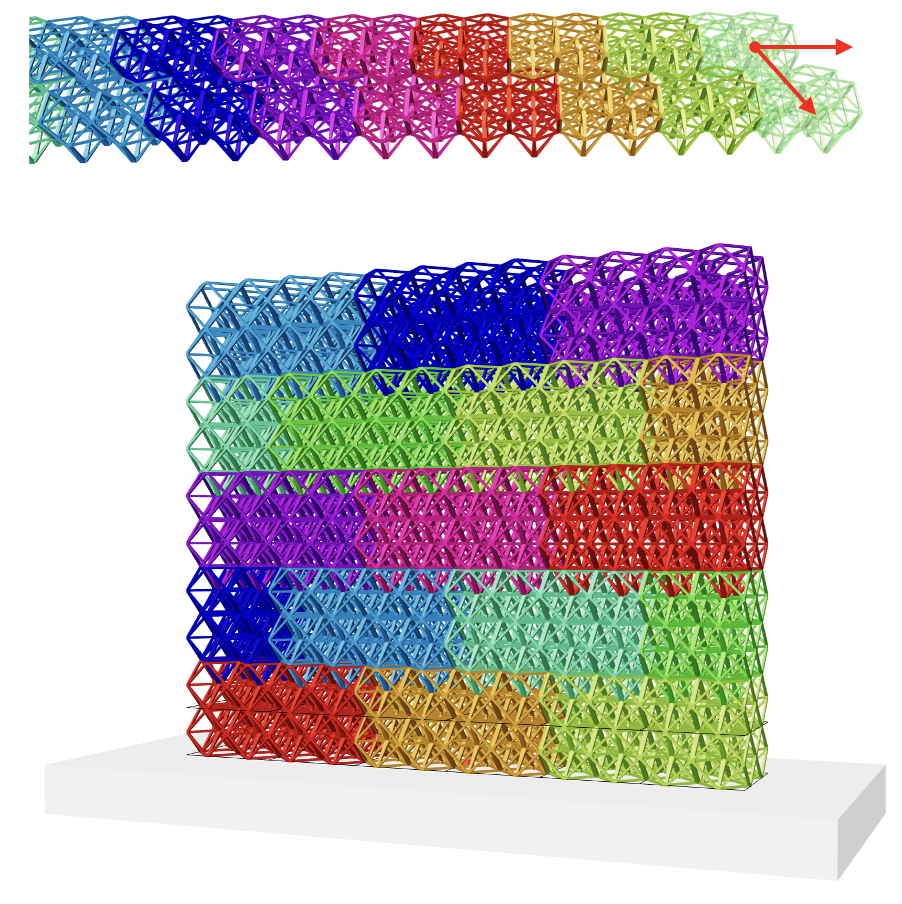}
  \caption{(Top) An example of tiling 2x2 offset voxel blocks to establish a first layer or overhang, with red arrows indicating the axes of potential extension. (Bottom) Beyond the first layer, layers can achieve interconnection through staggering layers.}
  \Description{Rendering showing how compounded voxel blocks tile.}
  \label{fig:voxel-tiling}
\end{figure}

We present a system for robotically assembling hierarchical discrete lattices. While prior voxel assembly systems have focused on a cuboctahedron lattice (equivalent to a node-connected octet lattice), we instead use an edge connected octet lattice because it inherently creates stable alignment features within each unit cell of the lattice (see Fig.\ref{fig:lattice-types} for reference). By pre-connecting these unit cells laterally, we create compounded lattice/voxel blocks that can then tile to form stable 1D, 2D, or 3D structures (see Fig.\ref{fig:voxel-tiling} for example tilings). These compounded blocks are thus all built from the same 1×1×1 unit voxel, but may vary in size and orientation (e.g., 2×2, 2×3, or 2×4), allowing a structure to mix different voxel types to optimize for geometry, performance, or assembly speed. The basic geometry of the lattice block is designed to self-align and constrain all but one degree of freedom when placed. In this work, the final degree of freedom is constrained with a releasable snap-fit (see Fig.\ref{fig:voxels-printed} for further detail). 

The self-aligning compounded voxel block offers a few key advantages for robotic fabrication: 1) the alignment features permit placement error on the order of 1/2 the lattice pitch; 2) the installation of new voxels is purely vertical; and 3) the snap fit connection does not require physical access to the connection plane to engage, but is still reversible. Together, these features substantially reduce the requirements on the robotic system. To successfully build up a structure, the robots only need to traverse over the existing lattice, carry new voxels, and coarsely place them. To this end, we developed inchworm-style robotic arms that can crawl over the structure, extending on the paradigm used in e.g. \cite{jenett_bill-e_2017}, \cite{park_soll-e_2023}, or \cite{abdel-rahman_self-replicating_2022}. 

However, the use of compounded blocks introduces new challenges to the design and path planning of these structures. To this end, we developed a new pipeline for intaking standard 3D meshes, voxelizing them, grouping the voxels into compounded blocks, and then simulating their robotic assembly. The simulation space is integrated with the hardware, so that it additionally acts as the central control/coordination for assembly. This workflow is shown in Fig.\ref{fig:teaser}. The software system is further discussed in Section 4, while the hardware system is discussed in Section 5, with assembly examples and system evaluation in Section 6.

%%%%%%%%%%%%%%
%%%%%%%%%%%%%%
\section{Software Implementation}
\sloppy
A custom Web-based software environment was developed in JavaScript, enabling full control and simulation of the robotic assembly pipeline on any internet-connected device, without requiring local installation. Leveraging the Three.js visualization library, the system offers real-time 3D interaction through a synchronized digital twin of the robot. 
The pipeline begins with voxelization of a 3D mesh and proceeds through build sequencing, path planning, and real-time visualization with robot feedback.

\subsection{Voxelization}

The first step in the pipeline is the voxelization of the input geometry, any 3D mesh in STL format. The mesh is discretized using the size of the unit voxel, resulting in a resolution of 65mm. Once voxelized, we analyze the grid to identify repeating patterns composed of multiple adjacent voxels. These patterns can vary in size and orientation and may be composed of simple or stacked configurations. Some example patterns are detailed in Figure \ref{fig:lattice-types}. To ensure connectivity and constructibility, the search prioritizes larger patterns first (e.g. 4x2x2 blocks) and proceeds hierarchically down to smaller units, filling in gaps as needed. Each detected pattern must be locally connected to at least one neighbor, either directly or via a base layer.

\subsection{Building Sequence}
\label{sec:building-sequence}
Given the location and type of each voxel in the structure, the next step is to define a build order that ensures feasibility while minimizing future path planning complexity. The building sequence determines the order in which each voxel of the structure is assembled and is influenced primarily by two parameters: the number of robots and the location of their associated voxel Feed. We assume each robot is assigned a unique voxel Feeder from which it collects and places voxels. 
The structure is first partitioned by assigning each voxel to the closest source based on the Manhattan distance. In case of equidistant voxels (a tie), a simple alternating policy ensures balanced distribution among Feeds.

Once assigned, the construction process is parallelized between robots. For each robot, a building sequence is computed independently. The structure is decomposed into horizontal layers, starting from the base and progressing upward. Within each layer, voxels are placed outward from the feed point, ensuring that the built region grows in a connected and stable manner. 

This strategy guarantees that every newly placed voxel rests on an existing one from the layer below, thereby creating a new accessible surface on the current layer. As a result, the robot can step onto this newly created platform to continue placing subsequent voxels. The building sequence algorithm enforces this outward growth, ensuring that at every stage, the next voxel to be placed is both structurally supported and physically reachable by the robot.

\subsection{Path Planning}
Once the build sequence is established, each robot must autonomously navigate from its associated Feed location to its assigned placement position, taking into account the current state of the partially build structure. To generate feasible and collision-free paths, we employ an A* algorithm adapted to the discrete 3D voxel space, which minimizes the function:
\begin{equation}
    h(n) = f(n) + g(n)
\end{equation}
Where g(n) is the cost of moving from the start to voxel n and h(n) is a heuristic estimating the cost to reach the goal from n. We use the Manhattan distance as the heuristic, which performs efficiently in the voxel grid while preserving path optimality in our constrained setup. 

The A* algorithm is particularly powerful in a voxelized environment, as the search space is inherently discrete, the connectivity is regular, and the heuristic directly reflects the grid geometry. By expanding nodes along the most promising directions first, it efficiently balances exploration and optimality. As a result, the algorithm consistently returns near-optimal paths with limited computation, even as the structure grows. Its effectiveness has already been validated in earlier iterations of this project \cite{abdel-rahman_self-replicating_2022, smith_self-reconfigurable_2024}.

\subsection{Robotic Assembly Simulation}

\subsubsection{Digital Twin for Live Simulation:}

To monitor and validate the robotic assembly process in real time, we developed a digital twin environment that mirrors the simulated robot's actions on the physical hardware. Path planning outputs are executed in the simulation and simultaneously relayed to the robot via a Python middleware layer, using WebSocket communication over Wi-Fi. This architecture offers several key benefits:
\begin{enumerate}
    \item Allows real-time updates during construction, enabling small-scale corrections in position or trajectory.
    \item Provides live mapping of the robot's state and the evolving structure.
    \item Offloads computation entirely to the simulation side, reducing the computational resources needed on the robot side.
\end{enumerate}
This tool is used within the pipeline to align the simulation with the hardware using the data flow presented in \ref{q}, with all elements generated directly from the target structure. Nevertheless, it can also be used as a standalone application, allowing the user to manually place every element of the simulation (voxelized environment, robots, feeds, voxels to build) and subsequently link the configuration to the hardware. An example is presented in Figure \ref{fig:Data Workflow}.

\begin{figure}[h]
  \centering
  \includegraphics[width=\linewidth]{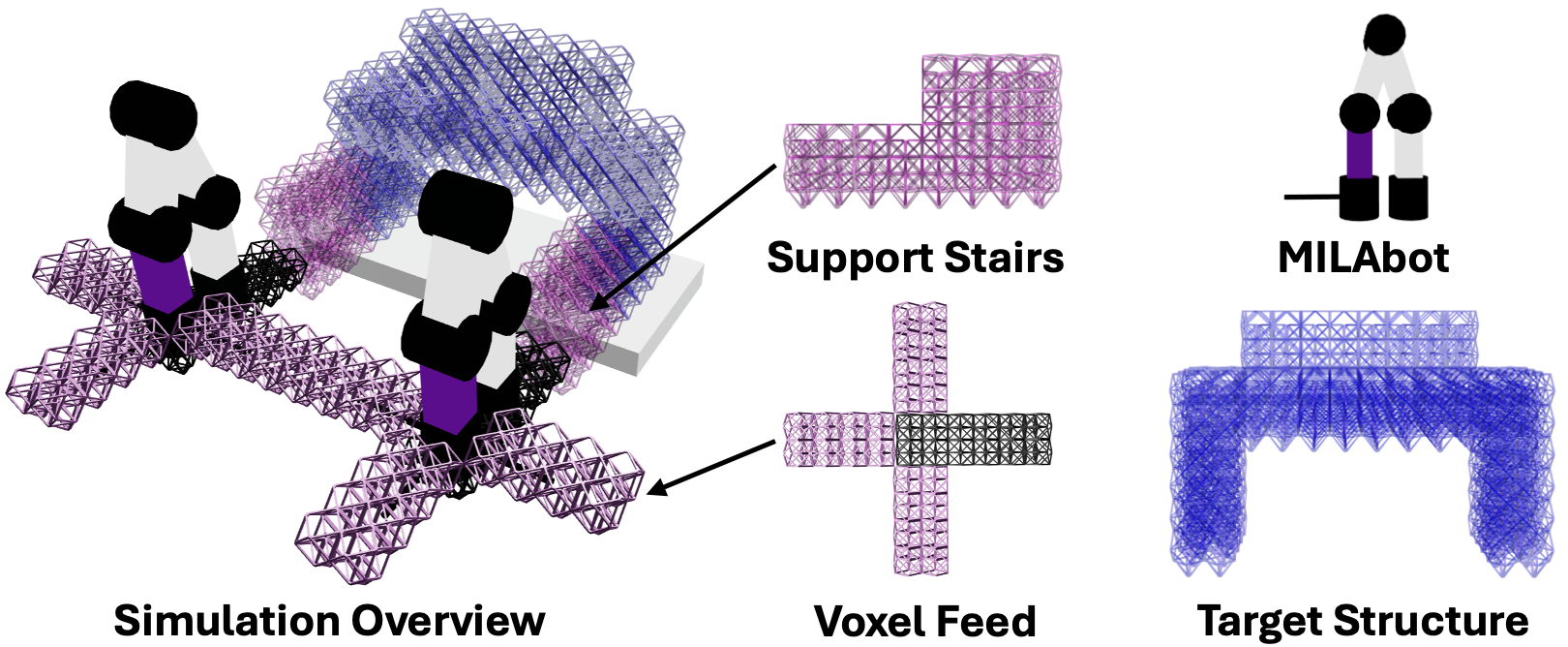}
  \caption{Simulation overview illustrating the four main elements: MILAbot, support stairs, voxel feed, and target structure.
}
  \Description{Graphical overview of lattice block types.}
  \label{fig:Simulation illustration}
\end{figure}

\subsubsection{Data Flow and Feedback:}
\label{q}
\begin{figure}[h]
  \centering
  \includegraphics[width=\linewidth]{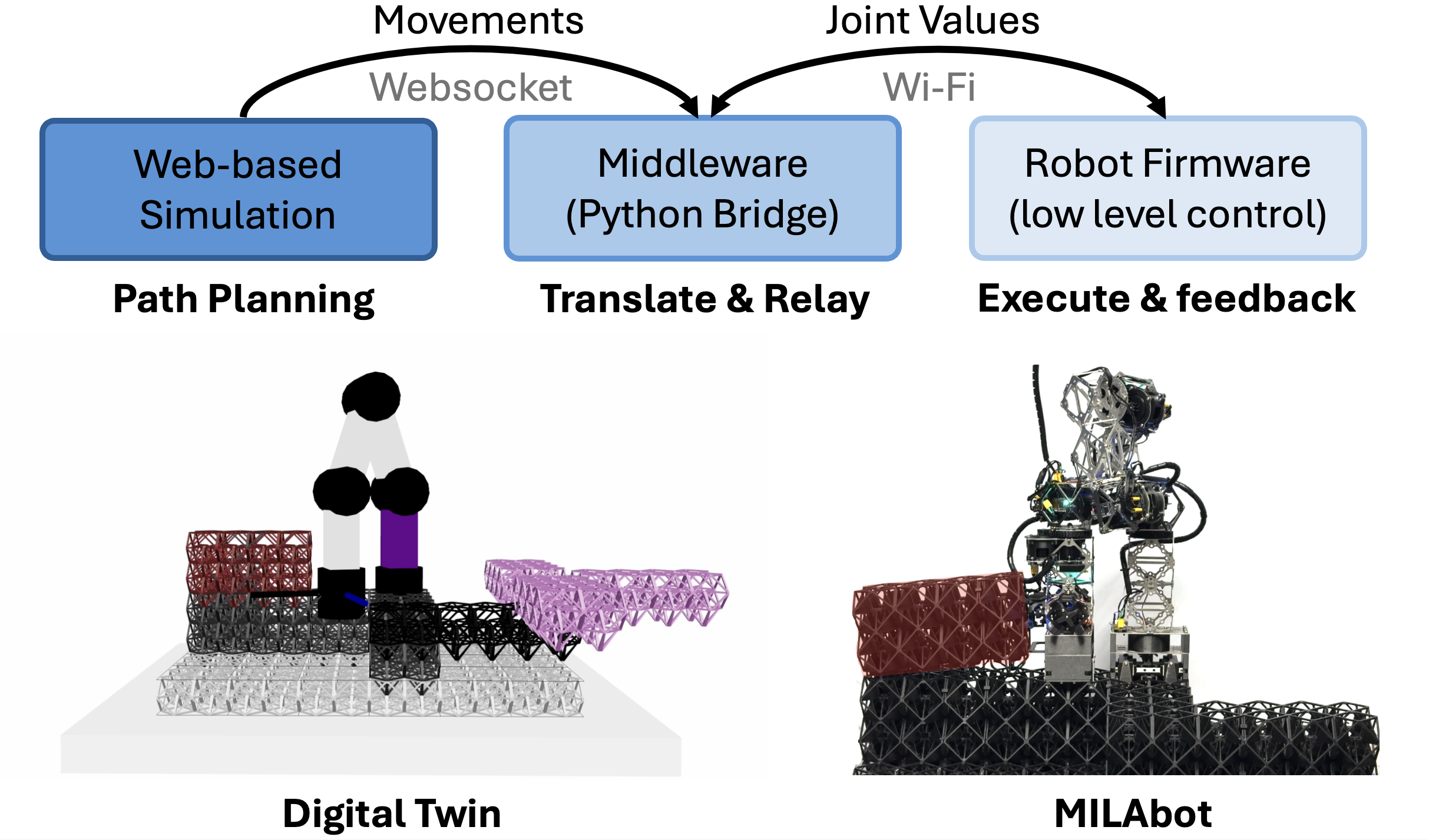}
  \caption{Top: Data flow from web-based simulation to MILAbot through middleware. Bottom: Digital twin synchronized with physical execution.}
  \Description{Workflow of the data in the simulation.}
  \label{fig:Data Workflow}
\end{figure}

The system operates using a feed-forward architecture: voxelization, build sequencing, and path planning are executed sequentially within the simulation environment. The resulting movement instructions are transmitted from the digital twin to the physcial robot— the Modular Inchworm Lattice Assembler robot (MILAbot)— via a Python bridge, which converts high-level actions into low-level joint target values using WebSocket communication. Simultaneously, the robot streams back its current joint states (see Figure \ref{fig:Data Workflow}). If a deviation from the expected pose is detected, a realignment sequence is triggered. This involves returning to the last valid position and re-approaching the target slowly to ensure accurate alignment without requiring full re-planning.

%%%%%%%%%%%%
%%%%%%%%%%%

\section{Hardware Implementation}

In this section, we discuss the hardware implementation of the voxel assembly system, covering the voxel design, manufacturing, and performance, as well as the robotic system design and operation. 

\subsection{Material System}

Our material system uses self-aligning compounded octet lattice blocks to create interlocking structures. In this work, we produce these lattice blocks in polylactic acid (PLA) through FFF 3D printing. We use FFF 3D printing because of its ability to easily handle complex geometries at a low cost and a reasonable speed. However, we envision that for future versions of this system, the compounded blocks might be further broken down into separate voxels, such as in \cite{gregg_ultra-light_2018}, or individual faces, such as in \cite{jenett_discretely_2020}, \cite{smith_self-reconfigurable_2024}, or \cite{jenett_building_2018}, so that they can be mass-manufactured from higher performance materials, such as GFRP in \cite{jenett_discretely_2020}, CFRP in \cite{gregg_ultralight_2024} or aluminum in \cite{smith_voxel_2025}. 

\begin{figure}[h]
  \centering
  \includegraphics[width=\linewidth]{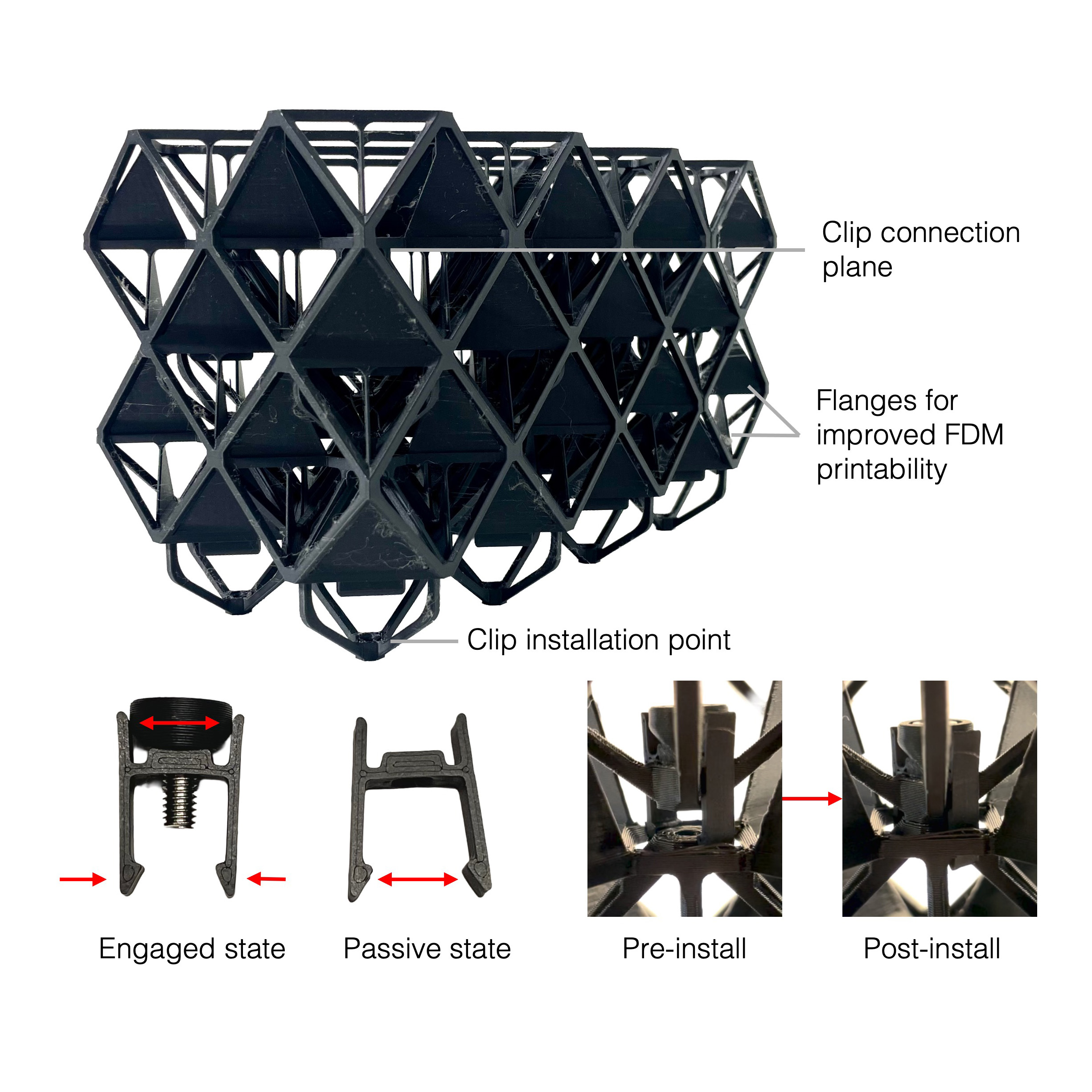}
  \caption{(Top) A 4x2x2 block of FFF printed PLA octet lattice with printability features added. (Bottom) Screw-release snap fit used for vertical connections.}
  \Description{Graphical overview of lattice block types.}
  \label{fig:voxels-printed}
\end{figure}

To improve the printability of the voxels, we add flanges to support the large overhanging regions during printing, as done in \cite{Leamon2025}. The flanges enable clean print quality without the use of supports, enabling more efficient material usage as well as improved absolute mechanical properties of the voxel block. However, the flanges are not intrinsically necessary, and it possible to print the lattice in its unmodified form. We printed the compounded voxels in different overall sizes depending on printer availabilty and print bed-size, and if necessary, later laterally pre-assemble them with printed snap clips. A representative 4x2x2 compounded block is shown in Fig.\ref{fig:voxels-printed}. 

Voxel blocks are vertically connected using a screw-released snap fit connector. The snap fit consists of three parts: a set of pincers that by default sit too wide to engage with a voxel below it, a screw (in this case an M4 8 mm long socket head screw) to fasten it directly into the base of a voxel, and a screw hat, which acts as a spacer to push the legs of snap pincers out, so that they can engage with the voxels below (these are shown in Fig.\ref{fig:voxels-printed}). The snap fit connectors are installed into the lowest octets of the printed voxels. The act of fastening the M4 screw then forces the pincers into the correct state, such that the voxel block can snap into one below it. To uninstall the connector, the M4 screw can be removed, which released the snap fit, allowing the block to removed. Though this is currently done manually, future versions of the robotic system could include an end effector for unscrewing the connectors. 

So that the voxels sit on a flat surface, we additionally produced base-layer voxel blocks that consist of the upper half of the standard block. These are added onto the protruding octets to create a flat base layer without adding extra height to the system. 

% \subsubsection{Mechanical Performance}

\subsection{Assembly Robots}

We introduce the Modular Inchworm Lattice Assembler robot, or MILAbot. As in prior voxel assembly robot systems, such as \cite{smith_self-reconfigurable_2024}, \cite{park_soll-e_2023}, \cite{jenett_materialrobot_2019}, the primary objective of the MILAbot design is to enable scalable voxel assembly with a system that can locomote over and assemble load-bearing discrete lattice structures. Looking toward scaling to the assembly of larger structures, the MILAbot is designed to increase assembly throughput while minimizing overall system cost and complexity. Additionally, the MILAbot is designed with eventual self-assembly in mind, that is, a starter MILAbot should be able to assemble more MILAbots, to autonomously improve efficiency as described in \cite{abdel-rahman_self-replicating_2022} and \cite{smith_recursive_2023}. 

\begin{figure}[h]
  \centering
  \includegraphics[width=\linewidth]{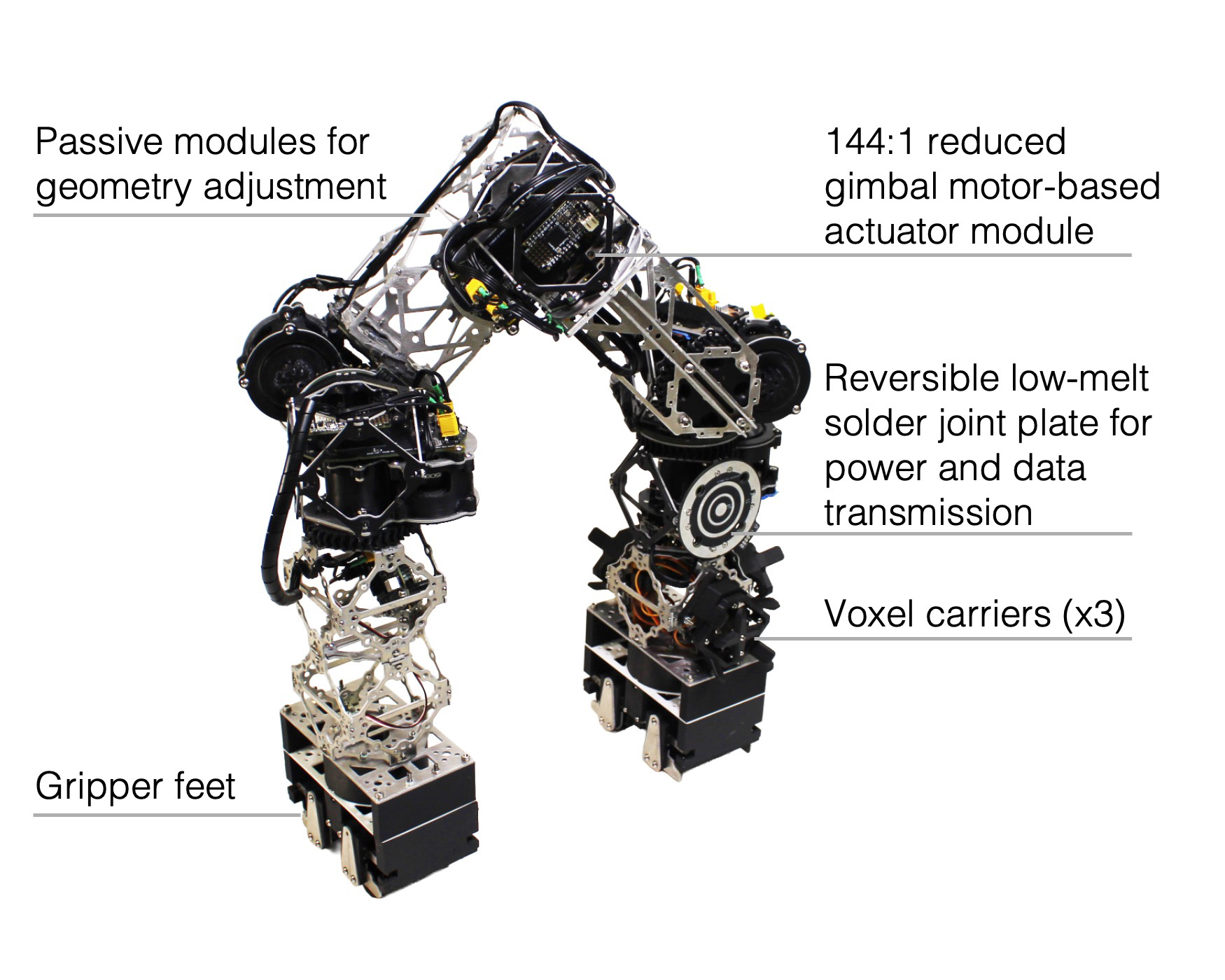}
  \caption{The Modular Inchworm Lattice Assembler robot (MILAbot) consists of four primary module types: actuated joints, passive spacers, gripper feet, and voxel carriers.}
  \Description{An image of the assembler robot (MILAbot) with primary components labeled.}
  \label{fig:robot-label}
\end{figure}

The MILAbot is made from four primary module types: actuated joints, passive spacers, gripper feet, and voxel carriers as showm in Fig.\ref{fig:robot-label}. These modules are arranged in a configuration similar to a five degree of freedom robot arm, with grippers on both ends, and three voxel carriers on one side. Toward eventual robotic-self assembly, the modules are capped by incoming- and outgoing- PCBs on either side that use a heating circuit to reflow low-melt solder between modules, forming a simultaneous eletrical and mechanical connection, such as in \cite{smith_self-reconfigurable_2024}, though this is not a focus of this work. The robot modules each have their own microcontroller, which are all networked over an I2C bus \cite{noauthor_i2c-bus_2021}, with a WiFi-enabled microcontroller acting as the primary microcontroller on the bus. We typically run the robot at 14V, but the electronics are designed to handle up to 24V. 

\begin{figure}[h]
  \centering
  \includegraphics[width=\linewidth]{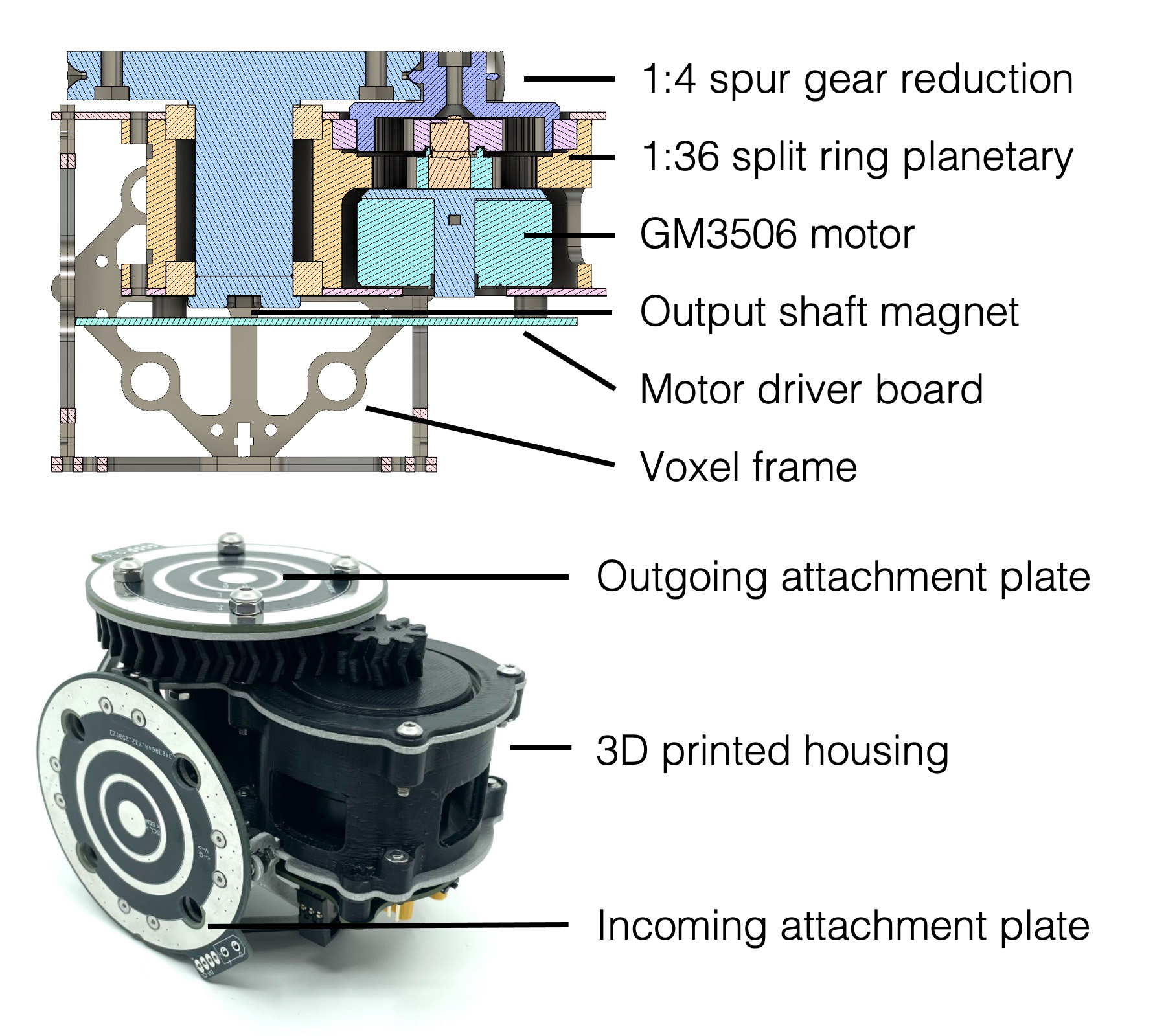}
  \caption{(Top) A cutaway view of the motor module design, which is based on a BLDC motor with two stages of reduction, feedback on the output shaft, and a voxel frame to enable module manipulation by another robot. (Bottom) A complete motor module, with incoming and outgoing PCB attachment plates.}
  \Description{A cutaway view of a MILAbot motor module and a photograph of a complete module, with parts labeled.}
  \label{fig:robot-module}
\end{figure}

The actuator modules are built around a GM3506 gimbal motor, a small brushless DC motor. We use a two stage 3D printed reduction consisting of a 1:36 split ring planetary followed by a 1:4 spur gear reduction, which has the added benefit of moving the output shaft off from the motor shaft, allowing us to more easily track the position of the output shaft. The total reduction is then 1:144. The module is fabricated from a mix of 3D printed polycarbonate, for its temperature resistance, and aluminum plates. The module controller board uses an Adafruit M4 Express (SAMD51) as the micronctroller, a DRV8316 for motor control, and two AS5407 magnetic encoders for motor and output shaft feedback. Motor control is done using SimpleFOC \cite{simplefoc2022}. Each module weighs approximately 400g and can output approximately 10-15 Nm before failure. 

\begin{figure}[h]
  \centering
  \includegraphics[width=\linewidth]{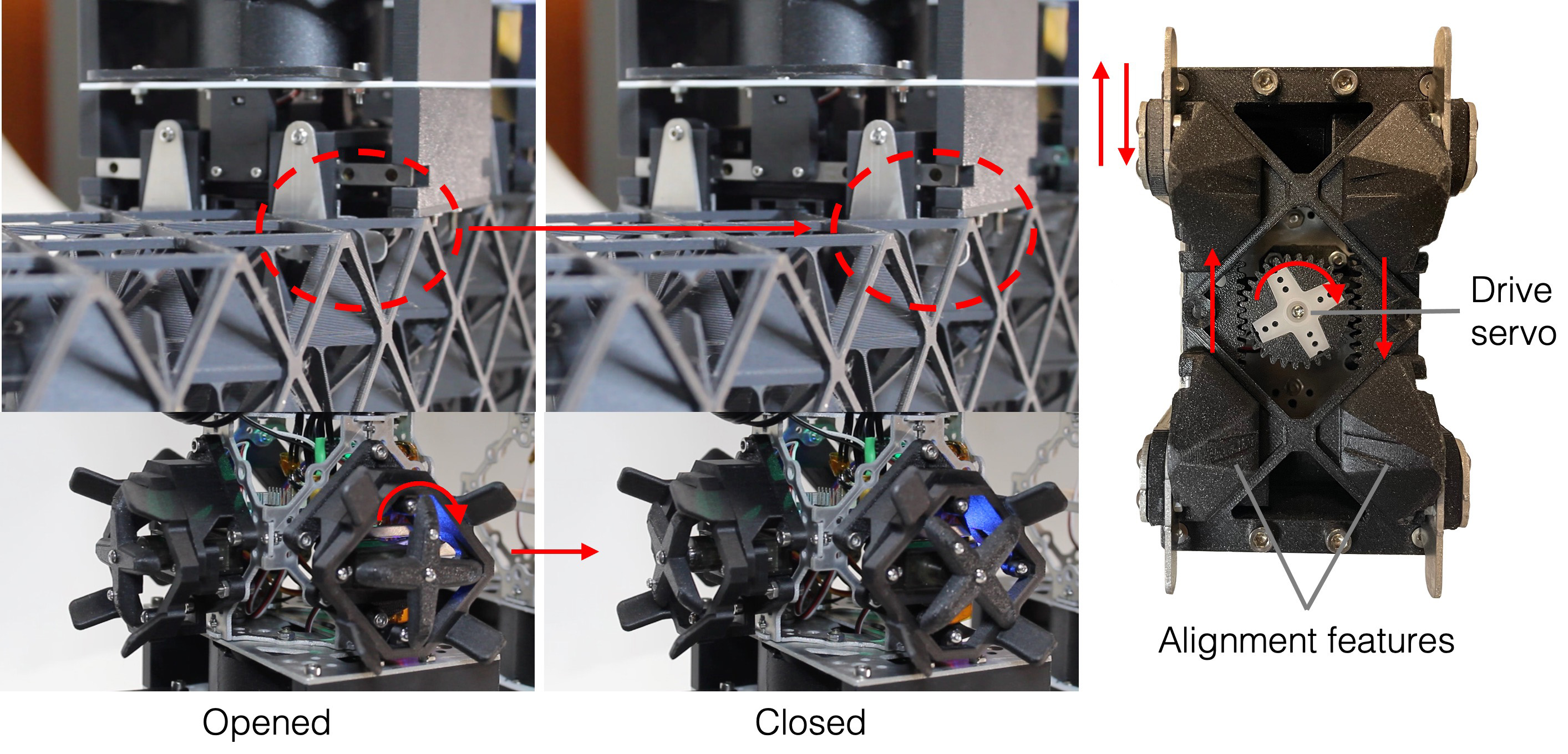}
  \caption{The operation of both gripper types in the MILAbot. (Top) The structure gripper, (bottom) the voxel payload gripper(s).}
  \Description{Photos of the voxel grippers.}
  \label{fig:grippers}
\end{figure}

The MILAbot uses two voxel gripper types. The first type adheres the robot to the structure, and the second type carries a voxel payload. The operation of both types is shown in Fig.\ref{fig:grippers}. The gripper feet are designed to step in the middle of a 2x2x1 grid of voxels, which introduces some constraints on where the robot can locomote. Its relatively larger size affords the robot greater stability and helps to distribute the load of the robot's locomotion over a larger area on the voxels. Each gripper foot is actuated using a FeeTech FS117 hobby servo, which retracts and extends a set of aluminum gripping features underneath the upper corner nodes of some voxels. 

Because the voxel carrier grippers do not need to support a significant load, they are designed to be smaller and lighter, using Miuzei MG90s micro servos to open and close a petal feature, modeled after the grippers used in \cite{jenett_materialrobot_2019} and \cite{abdel-rahman_self-replicating_2022}. To maximize the carrying capacity of the robot without significantly impacting its maneuverability, we use three per robot installed on one leg (though the robot could physically support installing an additional three voxel carriers on the other leg, this would cause significant collision issues for any non-flat structure). 

\begin{figure*}[h]
  \centering
  \includegraphics[width=\textwidth]{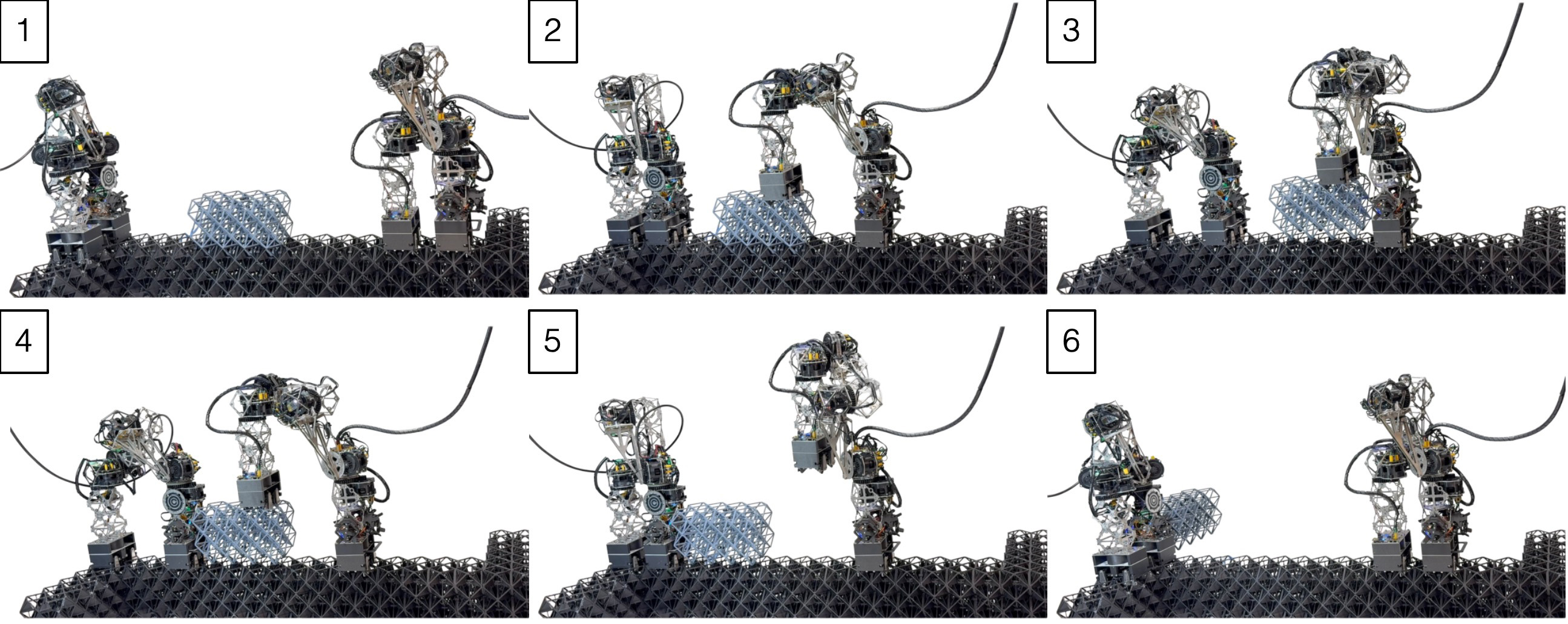}
  \caption{Picking up and passing a voxel block from one robot to another.}
  \Description{Freeze frames from a video of a robot handing a voxel block off to another robot.}
  \label{fig:voxel-pass}
\end{figure*}

Both grippers are designed with large alignment features. In the case of the gripper feet, these match the alignment features of the voxel blocks themselves, while the voxel carriers use a smaller petal shape. The combination of these mechanical alignment features in the robot and in the underlying structure means that we are able to achieve relatively precise robot and voxel placement without high-performance robots (i.e. the robots have minimal feedback and a lot of compliance). 

An example of this is shown in Fig.\ref{fig:voxel-pass}, where two robots pass a voxel block to each other. The two robots approach the target voxel (gray), and one steps onto it and lifts to remove it (the voxel is not constrained with snaps in this case). The other robot then steps forward, re-indexing its position on the lattice. Because the voxel carriers are placed on the robot at a multiple of the lattice pitch, this places the second robot at a "known" location for the robot carrying the other voxel, which only needs to then move into that position, with the alignment petals proving fine alignment. The second robot then grips the voxel while the first lets go and retracts.

\begin{figure}[h]
  \centering
  \includegraphics[width=\linewidth]{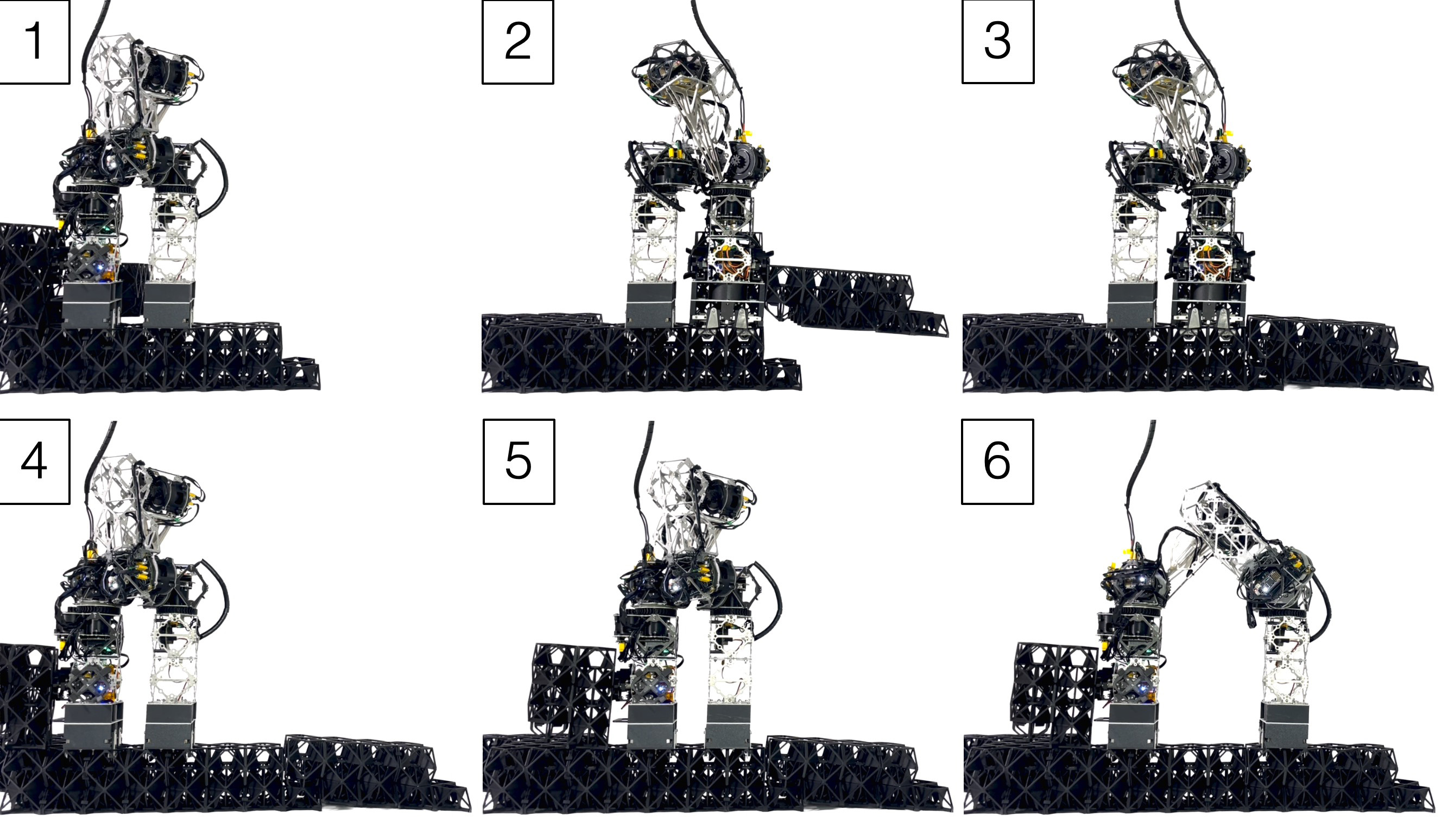}
  \caption{An example voxel installation sequence. 1) The robot is in pre-placement position. 2) The robot steps into the placement position and 3) drops the voxel roughly into place, before 4) retracting, and then 5) taking a step forward. 6) The robot stomps the new voxel to engage the snap fit connection.}
  \Description{Freeze frames from a video of a robot installing a voxel.}
  \label{fig:voxel-install}
\end{figure}

The robot relies on the ability of the voxels to correct large placement errors to install new blocks simply by dropping them into place and then stepping on them to engage the snap fits. This process is shown in Fig. \ref{fig:voxel-install}, where a robot carrying two voxels steps into position, swings the back leg forward and rotates it to place the correct voxel in the appropriate pre-placement position, drops the voxel, returns the back leg backwards, steps forward once, and then stomps on the voxel to engage it, forming a sturdy connection. 

\section{Evaluation}

In this section, we evaluate the performance of the presented system through a set of assembly demonstrations, the throughput and fidelity of the mechanical system, the mechanical efficiency of the lattice type, and general robotic performance metrics. 

\subsection{Assembly Demonstrations}

We use the complete workflow to demonstrate robotic assembly of basic structures using one and two robots. 

\subsubsection{Single Robot Assembly}

\begin{figure}[h]
  \centering
  \includegraphics[width=\linewidth]{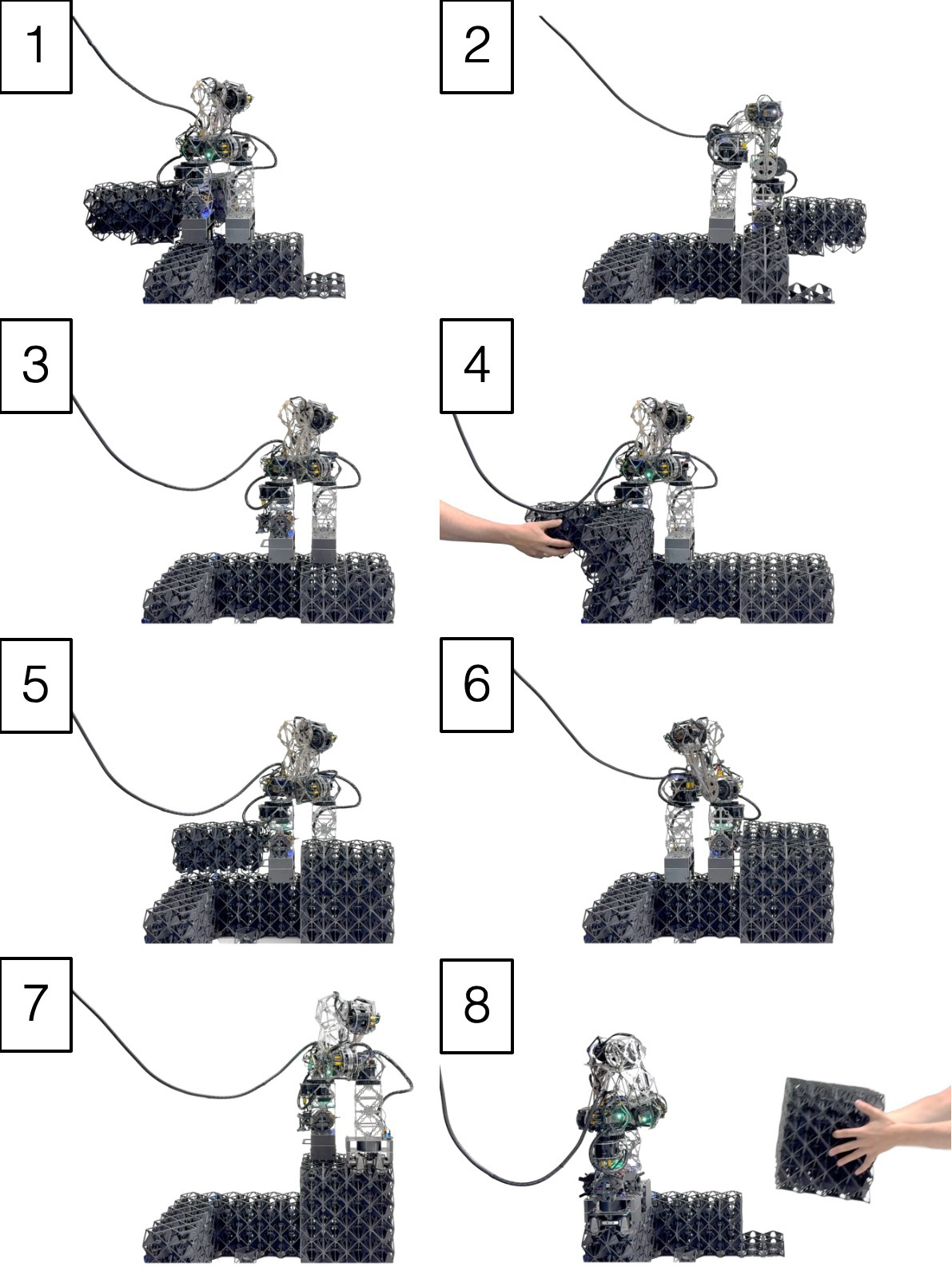}
  \caption{Assembly of a 4x4x4 voxel block using a set of four compounded blocks, carried two at a time.}
  \Description{Freeze frames from a video of a robot building a voxel cube.}
  \label{fig:cube-build}
\end{figure}

First, we use a single robot to assemble a 4x4x4 voxel block. We voxelize the structure into four 4x2x2 blocks, the largest compounded size our system currently accommodates. This way, the robot only needs make four placements, in contrast to prior approaches to voxel assembly, which would have required 64 voxel placements for the equivalent structure. A condensed version of the assembly sequence is shown in Fig. \ref{fig:cube-build}. The robot carries two blocks at once, as the minimum amount of trips it can make for this structure is two (either two voxels twice, or three voxels and one voxel once). The first two voxels are placed oriented toward the camera plane, while the second two are placed parallel to it, resulting in a fully connected structure.

\begin{figure}[h]
  \centering
  \includegraphics[width=\linewidth]{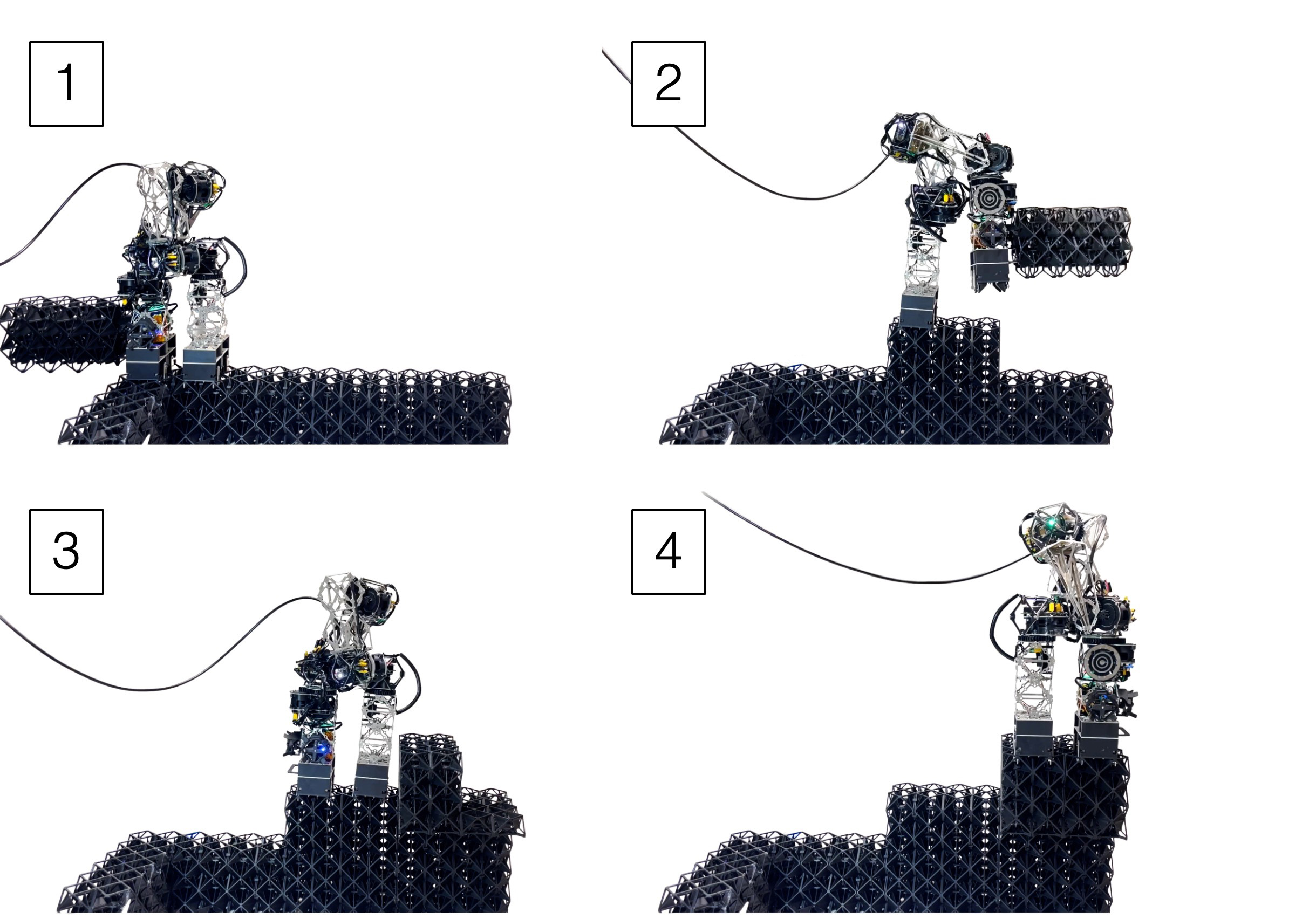}
  \caption{A single robot assembling a support staircase for the bench build.}
  \Description{Freeze frames from a video of a robot building a voxel staircase.}
  \label{fig:stairs-build}
\end{figure}

Next,  each robot assembles its support staircase for the bench example, as previously shown in Fig.\ref{fig:teaser} and Fig.\ref{fig:Simulation illustration}. Freeze frames from the build sequence are shown in Fig.\ref{fig:stairs-build}. The build process for the stair case is similar to that of the cube, but at the end, we demonstrate a small overhang with this structure that is able to support the weight of the assembler robot on it. Overhangs, or first layers, both require staggered voxel blocks to achieve. This can be done either by layer shifting a 2-voxel tall stack, or by attaching base-plate voxels at a stagger, as done here. The offset provides alignment features for the next block placement and braces the newly installed voxel against the existing structure, especially during the drop-to-place procedure. 

\subsubsection{Double Robot Assembly}

Because each robot has its own voxel feed in our workflow, most multi-robot structure assembly is identical to single robot assembly, up until the point at which the structures meet. The joining of separate build fronts, or unlinked structures, is not something considered by any of the prior voxel assembly systems, but is likely a critical feature for scaling these systems up. 

\begin{figure*}
  \centering
  \includegraphics[width=\textwidth]{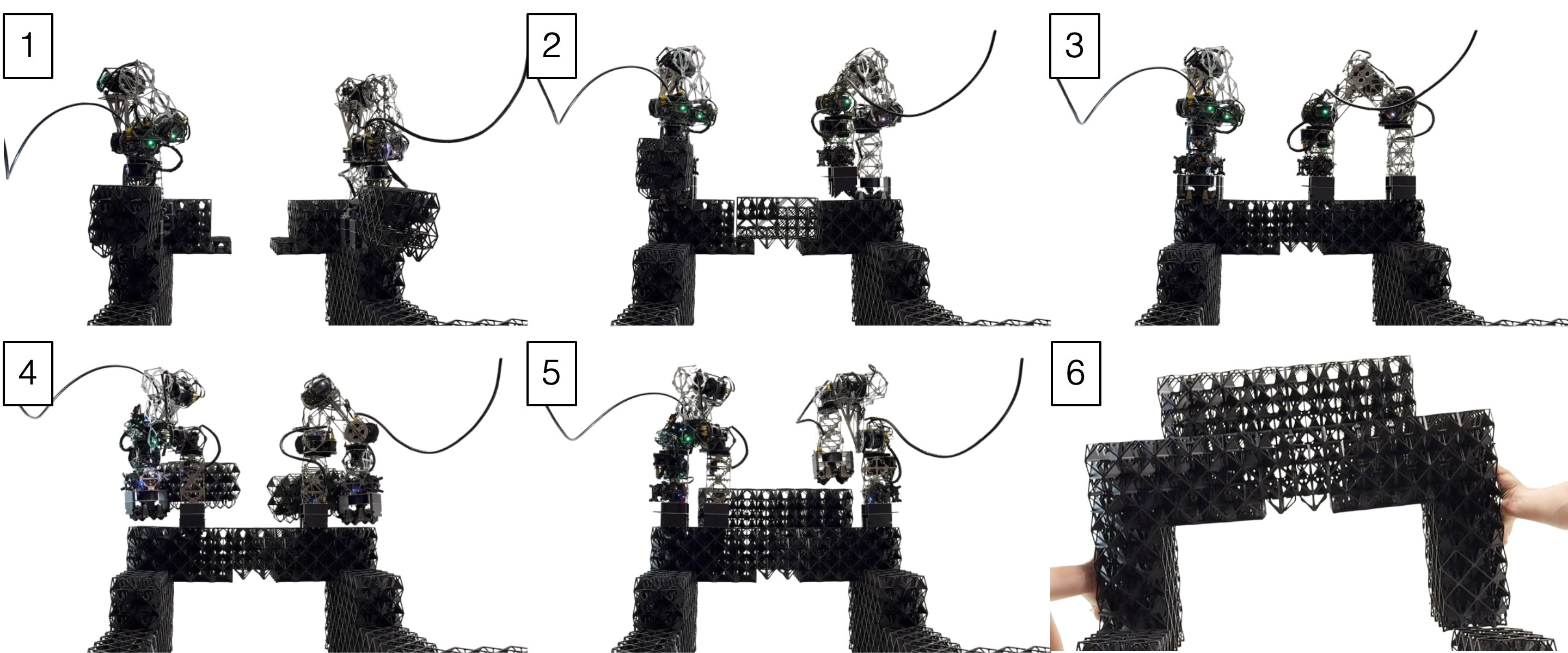}
  \caption{Two robots assembling a bench together.}
  \Description{Freeze frames from a video of a robot building a voxel bench.}
  \label{fig:bench-build}
\end{figure*}

In this example, the two robots finish the assembly of the bench by filling in the base of the seat (see Fig.\ref{fig:bench-build} for reference). One robot drops a voxel block into the empty space, and then steps on it to align it and the structure. The second robot then installs its voxel into the remaining space. This is possible because the alignment features on the voxels are permissive of a fairly large amount of error, the error is larger than the voxels, and the structure is able to slightly slide and deform. We cannot always guarantee that this will be the case, so for joining build fronts, it is likely necessary to consider alternative voxel types, such as ones that are slightly undersized or compliant, to account for the potential misfit. However, even though this is not a reliable method for joining a voxel seam, this block placement is taken to be infeasible in the prior planning systems for voxel assembly \cite{gregg_ultralight_2024}, \cite{smith_self-reconfigurable_2024}, \cite{jenett_materialrobot_2019}. 

After installing the base of the bench, the two robots install the back of the seat. These two voxels are necessarily installed simultaneously, as to avoid collisions in the placement. Though for a small structure with only two robots, keeping synchronicity (or restoring it if lost) is not difficult, as the system scales to larger robot counts, it may be necessary to consider additional methods for synchronizing the movements of all robots at critical junctures, without needing to run the entire system at the pace of the slowest robot. Or, the methodology for voxel placement could be revisited, to consider more sophisticated obstacle avoidance, depending on the build state of the structure. 

\begin{figure}[h]
  \centering
  \includegraphics[width=\linewidth]{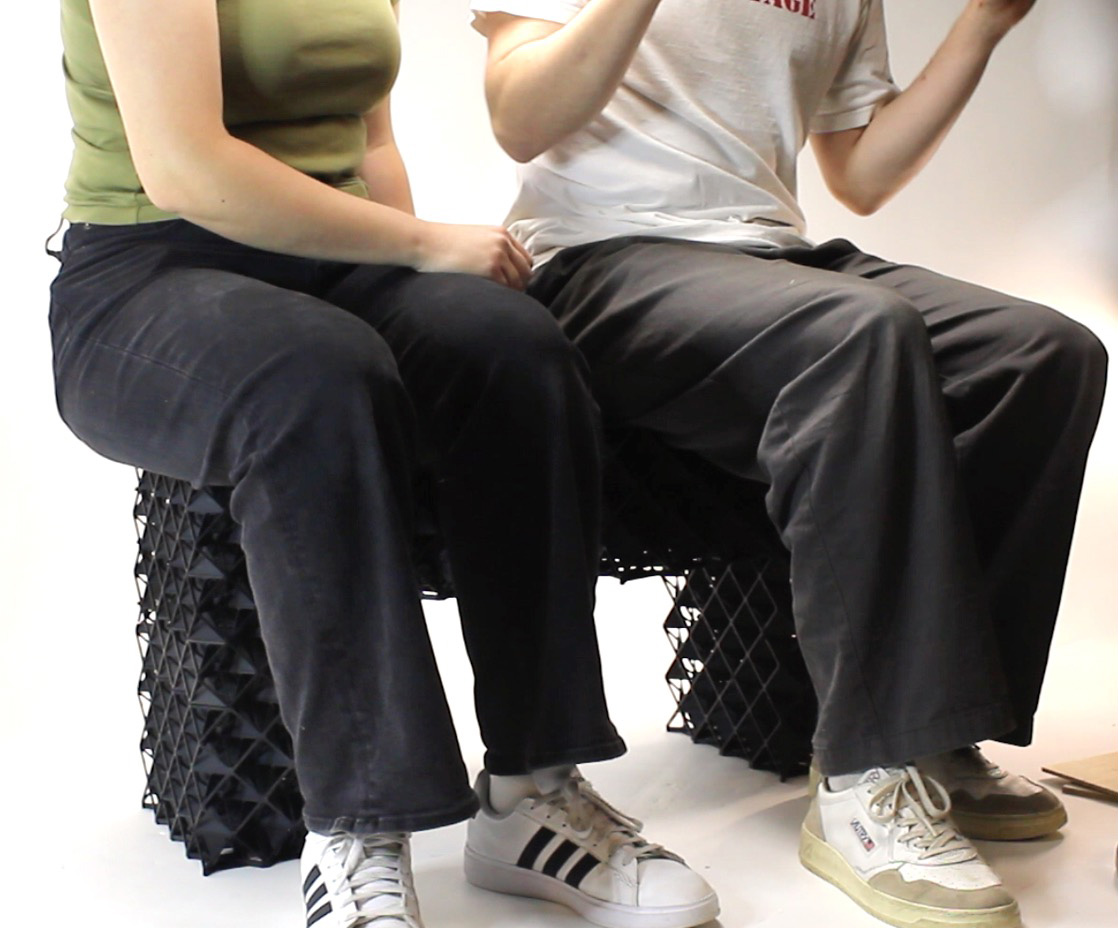}
  \caption{The assembled bench supports two people.}
  \Description{A photograph of two people sitting on the assembled voxel bench.}
  \label{fig:bench-sit}
\end{figure}

Once completed, the bench is removed from the support staircase, where the path planning has left a seam. The completed bench is then directly usable, as the voxels have decent mechanical performance, which is further discussed in Section 6.3. We qualitatively evaluated the bench by having two people sit on it, as shown in Fig.\ref{fig:bench-sit}. The bench supports this weight, though the overall design is a bit small for comfort— luckily, the voxels support disassembly and reassembly, so we can change the design. 

\subsection{Throughput and Fidelity}
\begin{figure}[h]
  \centering
  \includegraphics[width=\linewidth]{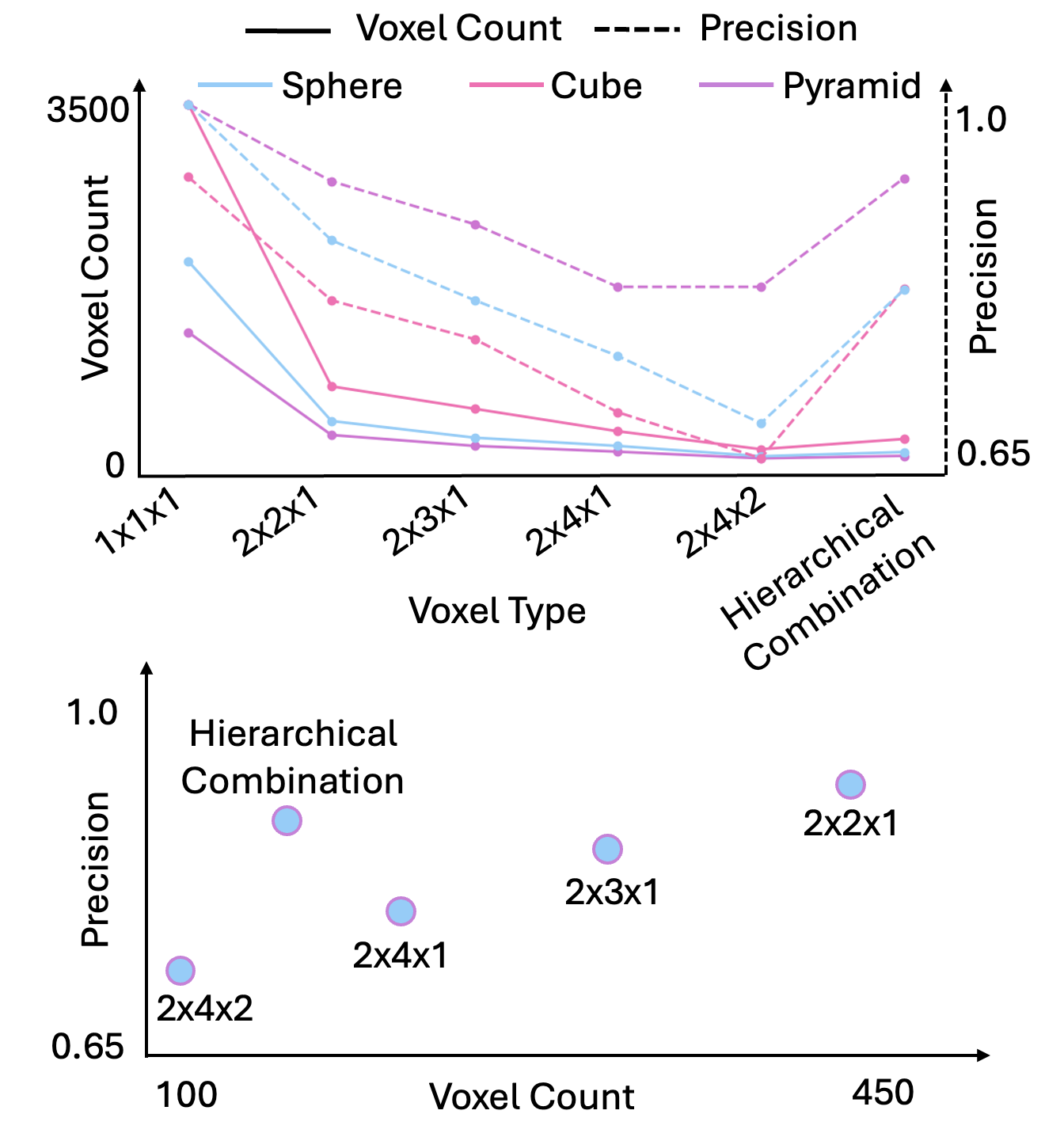}
  \caption{Evaluating voxel pattern efficiency—  hierarchical combination outperforms any single voxel type by breaking the linear trade-off between voxel size and precision. (Top: Voxel count and precision decrease linearly with increasing voxel size when voxelizing meter-scale meshes. Bottom: Precision as a function of voxel count, highlighting the benefits of the hierarchical combination.)}
  \Description{Voxelization Count and Precision vs Pattern Type for multiple meter scale mesh (Left). Precision vs Voxel Count for these same mesh(Right)}
  \label{fig:voxel-efficiency}
  
\end{figure}

As shown in Figure~\ref{fig:voxel-efficiency}, we assess the performance of different voxelization strategies by analyzing the trade-off between the number of voxels that make up the structure and geometric precision  across various mesh types. The geometric precision can be defined as:
\begin{equation}
    precision = n_{voxel} * V_{voxel} / V_{mesh}
\end{equation}

Where $n_{voxel}$ is the number of unit voxel with $V_{voxel}$ being the volume of one unit voxel and $V_{mesh}$ the volume of the input mesh.

The left panel shows how larger voxel patterns drastically reduce the number of elements required but come with a loss in precision. The right panel highlights this trade-off by plotting precision as a function of voxel count. Patterns like 2x3x1 and 2x2x1 achieve a good balance, while the hierarchical combination offers a compelling middle ground with high precision and low voxel count.

\begin{figure*}
  \centering
  \includegraphics[width=\textwidth]{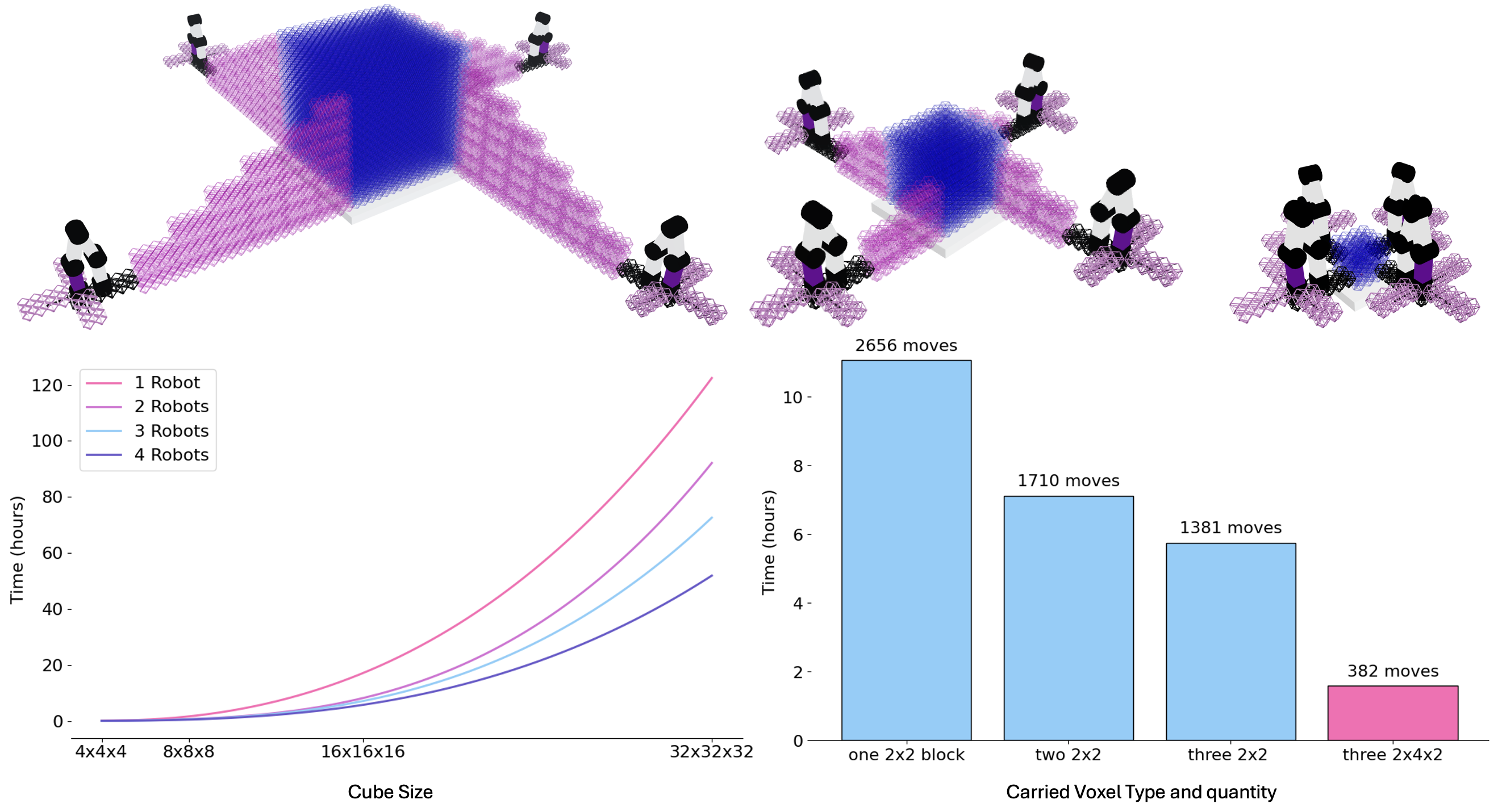}
  \caption{Impact of robot quantity and voxel carrying strategy on build time. (Top) Simulation snapshots of cubes of decreasing size (16x16x16, 8x8x8, and 4x4x4 voxels). (Bottom Left) Estimated build time as a function of cube size and number of robots. (Bottom Right) Time required to build the same structure depending on the number or size of voxels carried per trip. These graphs motivate the use of a larger robot fleet for assembling large structures, as they show a substantial decrease in assembly time. They also highlight the importance of maximizing robot payload, with each robot ideally carrying three 4×4×2 voxels rather than operating below full capacity.}
  \Description{Impact of robot quantity and voxel carrying strategy on build time. (Top) Simulation snapshots of the same structure being built by 1 to 4 robots. (Bottom Left) Estimated build time as a function of cube size and number of robots. (Bottom Right) Time required to build the same structure depending on the number or size of voxels carried per trip.}
  \label{fig:scaling-carrying-impact}
\end{figure*}

Figure~\ref{fig:scaling-carrying-impact} highlights two key components to accelerate construction: increasing the number of robots and improving the voxel carrying strategy. On the top row, simulation snapshots show how multiple robots parallelize the construction of a 16x16x16 cube, reducing total build time. While the benefit is clear, the speed-up is not exactly proportional to the robot quantity due to each robot building its own support stairs independently, without sharing intermediate infrastructure.

The graph on the bottom left shows that construction time increases exponentially with structure size, but adding robots significantly flattens the curve. On the bottom right, we analyze the impact of voxel carrying capacity. Carrying multiple small voxels per trip already provides notable gains, but the greatest efficiency comes from transporting larger compound blocks (e.g., 4x2x2). This confirms that both scaling the number of robots and optimizing carried voxel configuration are critical for improving overall build throughput.

\subsection{Hardware Performance}

Because other robotically assembled discrete lattices have been demonstrated, we can compare the performance of our system against those in literature. First, we consider the mechanical performance of the modified octet lattice we use. In evaluating this performance, we consider the version of the lattice presented here to be a proof-of-concept demonstration, and the comparisons below to indicate, generally if the design decisions we have made are in desirable directions. Because of the anisotropy in FFF printed parts, the robustness of our voxel system can be low, and we do not perform a full mechanical characterization of the voxels under tensile and bending loads. Further, the overall interplay between vertically interconnected blocks under various loads is also still unknown, and beyond the scope of the current work.

To determine the behavior of the lattice under compressive loads, we tested a single unit cell, a 2x2x2 lattice block, and a 3x3x3 lattice block on an Instron 4411 with a 5kN load cell for the single unit at the 2x2x2, and on an Instron 5985 with a 250kN load cell for the 3x3x3 at a compression rate of 10 mm/min.  The results of this testing are summarized in Table \ref{tab:mechanical_properties}. The lattices demonstrate good strength and stiffness at a reasonable weight (the average density of the lattice is 81.85 $\mathrm{kg/m^3}$). Under compression, the lattices can support very high loads— the 2x2x2 block reached a maximum load of 3445N, which is about 2,220x its own weight, while the 3x3x3 block reached a maximum load of 8712N, which is about the load exerted by the weight of an average cow. 

We can determine the mechanical efficiency of the lattice by looking at how its compressive modulus and density scale relative to the bulk material it is made from. The relative compressive modulus is given by $E / E_s$ where $E$ is the compressive modulus of the lattice and $E_s$ is the compressive modulus of the bulk material. Similarly, relative density is given as $\rho / \rho_s$, where $\rho$ is the density of the lattice and $\rho_s$ is the density of the bulk material. Ideal stretch dominated behavior, that is, the most efficient scaling of modulus with density, is given by linear scaling $E/\rho$. However, this is generally considered inaccessible \cite{schaedler_architected_2016}. Instead, ideal bending dominated behavior is given by quadratic scaling $E^{1/2}/\rho$, while below $E^{1/3}/\rho$ is accessible via foams. So, the target for the structural efficiency of an architected lattice is to surpass the quadratic scaling for its bulk material. 

We plot a comparison of the relative compressive modulus and relative density for this work, as well as the other discrete lattice assembly systems: \cite{gregg_ultralight_2024}, \cite{smith_self-reconfigurable_2024}, and \cite{jenett_materialrobot_2019}. Both this work and \cite{gregg_ultralight_2024} achieve stretch dominated behavior, indicating efficient and desirable material usage. Note that \cite{jenett_materialrobot_2019} does not provide mechanical testing data; this is instead estimated using an ideal beam simulation of the presented lattice using the reported parameters and materials. Additionally \cite{jenett_materialrobot_2019} uses a non-structural magnetic joint, but for this comparison we disregard that. 

We can additionally compare the assembly throughput and system cost for these systems. We use the reported volumetric assembly throughput for each system, with the exception of \cite{gregg_ultralight_2024}, in which their reported throughput is for a hundred voxel structure, and so is an underestimate of the speed of the system as compared to the systems in \cite{jenett_materialrobot_2019} and \cite{smith_self-reconfigurable_2024} which only demonstrate assembly of tens of voxels. Instead, we draw an average from the installation time per voxel for the first 15 voxels in their structure, to better align with the data we have here, resulting in a much higher throughput estimate than what they report. 

We find that our volumetric assembly throughput is 4,394,000 $\mathrm{mm^3/min}$ for a single robot carrying two blocks of 4x2x2 voxels, as compared to 2,700,000 $\mathrm{mm^3/min}$ for \cite{gregg_ultralight_2024}, 751,879 $\mathrm{mm^3/min}$ for \cite{jenett_materialrobot_2019}, and 343,281 $\mathrm{mm^3/min}$ for \cite{smith_self-reconfigurable_2024}. Our higher volumetric throughput is owed to two main factors: we can carry multiple blocks of compounded voxels while moving at a comparable speed to \cite{jenett_materialrobot_2019}, and unlike \cite{gregg_ultralight_2024} the current implementation requires minimal installation/fastening time. This result indicates that we likely have budget to increase the time spent per voxel (such as by implementing a different connection system) while still staying competitive in volumetric throughput. 

We additionally compare these results against the cost of each robotic system. For \cite{smith_self-reconfigurable_2024} and \cite{jenett_materialrobot_2019} these values are taken from their reported bills of materials, while for \cite{gregg_ultralight_2024} we estimate the cost based on the components reported in \cite{park_soll-e_2023}, where the current figure is based on doubling the cost of the actuation system for one inchworm robot, to account for the need for both a carrying robot and an installing robot, with an additional \$300 added as a low-end estimate for the cost of the third robot, as well as all of non-actuation components in the inchworm robots. Interestingly, there is a relatively linear relationship between system cost and assembly throughput for the prior single-voxel based systems, which this work sidesteps via implementing hierarchical material handling in an otherwise very simple robot.

\begin{table}[h]
    \caption{Summary of mechanical properties for different voxel block sizes}
    \centering
    \begin{tabular}{lccc}
    % \hline
    \toprule
    \textbf{} & \textbf{1x1x1} & \textbf{2x2x2} & \textbf{3x3x3} \\
    % \hline
    \midrule
    \textbf{Stiffness [N/mm]} & 949 & 2278 & 4868 \\
    \textbf{Maximum load [N]} & 602 & 3445 & 8712 \\
    \textbf{Compressive Modulus [MPa]} & 14.6 & 17.5 & 24.9 \\
    % \hline
    \bottomrule
    \end{tabular}
    \label{tab:mechanical_properties}
\end{table}

\begin{figure}[h]
  \centering
  \includegraphics[width=\linewidth]{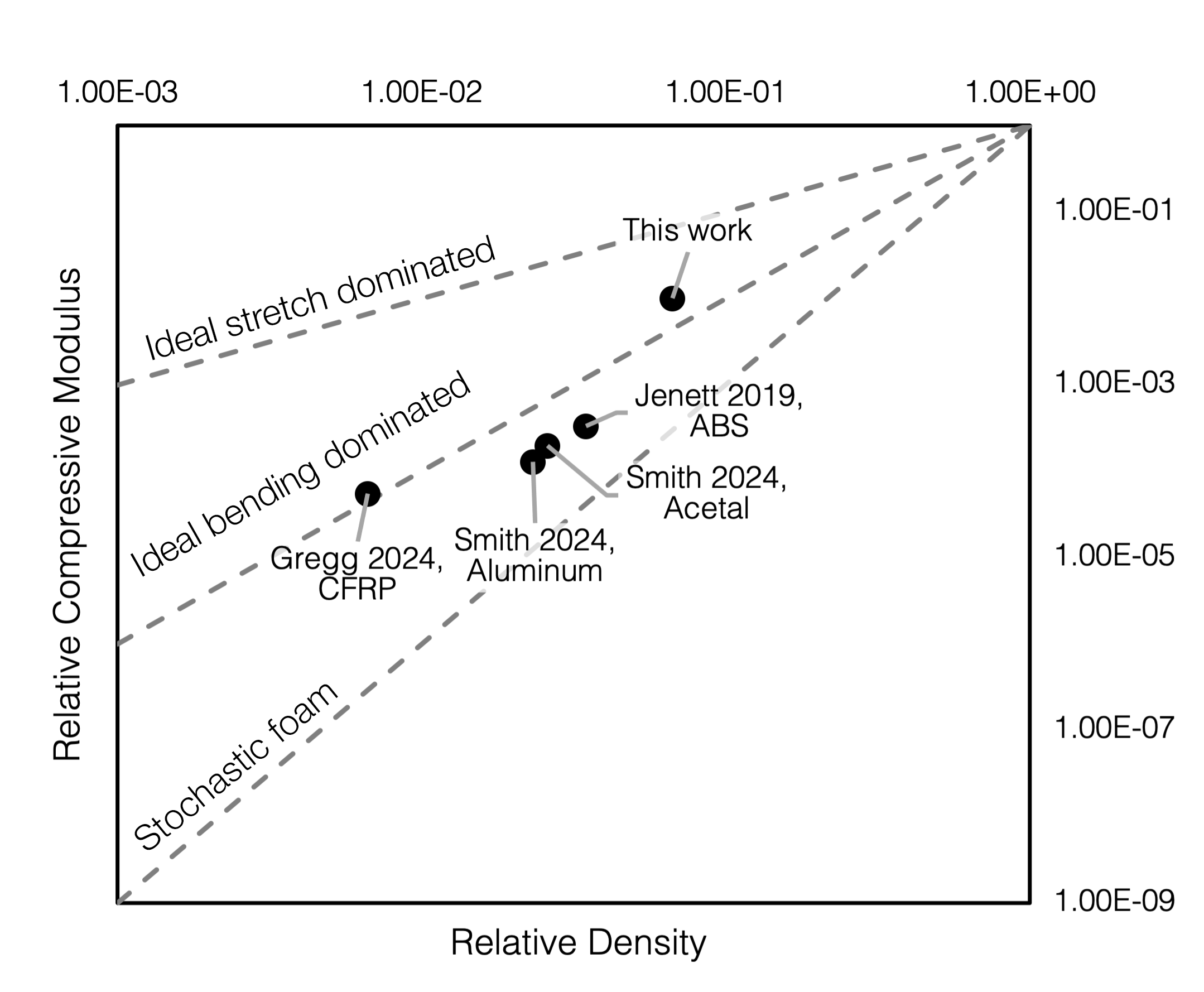}
  \caption{A comparison of the relative compressive modulus (voxel modulus divided by bulk material modulus) and the relative density (voxel density divided by bulk density) for robotically assembled discrete lattices. This work achieves stretch dominated behavior, demonstrating material efficiency relative prior robotically assembled voxel systems. }
  \Description{Graph of relative compressive modulus and relative density for robotically assembled lattices.}
  \label{fig:lattice-efficiency}
\end{figure}

\begin{figure}[h]
  \centering
  \includegraphics[width=\linewidth]{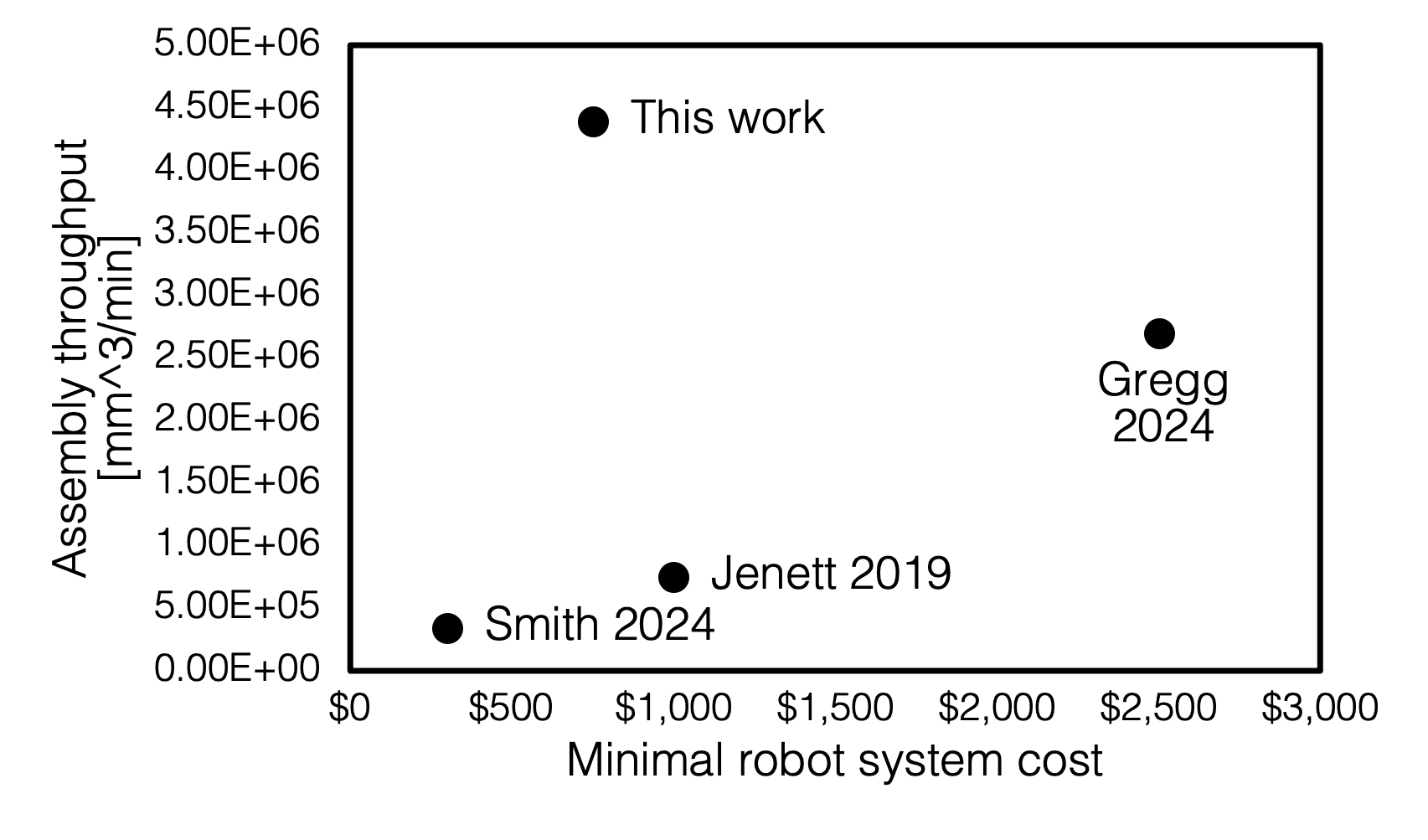}
  \caption{A comparison of the volumetric throughput for different voxel assembly systems for low voxel count structures, against the cost of the minimal robotic assembly unit. This work achieves the highest throughput at the second lowest cost.}
  \Description{Graph comparing volumetric throughput and cost for voxel assembly systems.}
  \label{fig:cost-throughput}
\end{figure}

\section{Future Work and Limitations}

\subsection{Voxelization}
\begin{figure}[H]
  \centering
  \includegraphics[width=\linewidth]{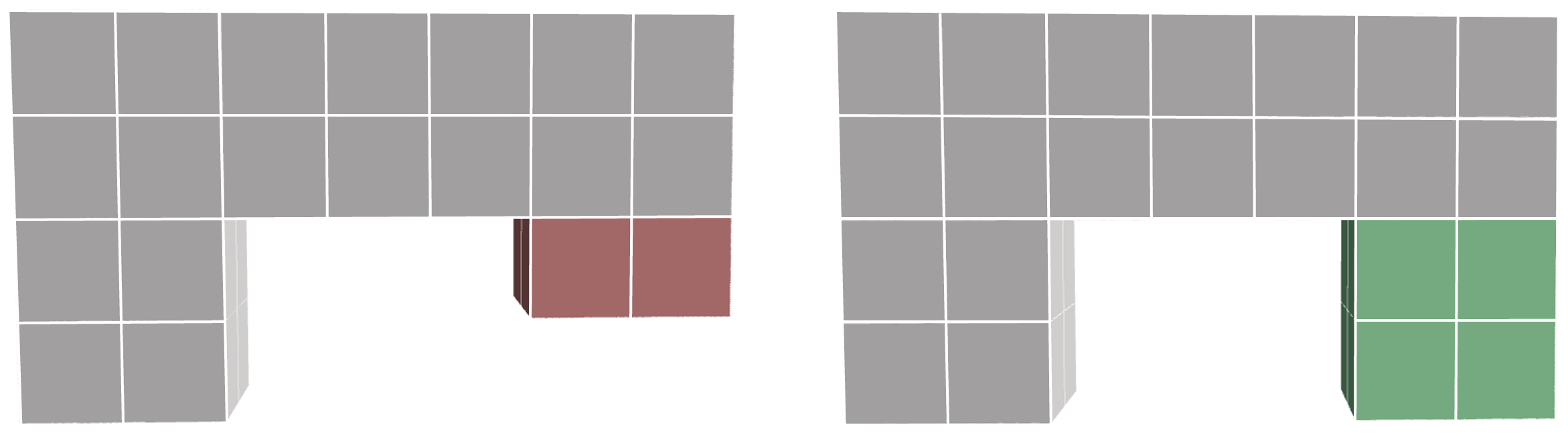}
  \caption{Comparison of infeasible (left) and feasible (right) structures due to local connectivity constraints.}
  \Description{figure showing what kind of structure are build-able.}
  \label{fig:possible structure}
\end{figure}

While our voxelization algorithm proved sufficient to demonstrate the feasibility of a complete robotic assembly pipeline across a defined set of structures, it remains limited in its generalizability and is sensitive to variations in input geometry. In future work, we plan to explore more robust and generalizable approaches. This includes developing enhanced rule-based algorithms inspired by an octree subdivision approach as developed in \cite{abdel-rahman_self-replicating_2022}, which remain fully discrete and deterministic, as well as investigating learning-based methods, such as those proposed in \cite{pun2025generatingphysicallystablebuildable}, that leverage the power of large language model to Generate physically stable and buildable LEGO® structures. Incorporating a structure-aware strategy that accounts for physical constraints and evolving geometry could significantly improve the reliability and feasibility of voxel placement over time. And would extend the buildable structure to the one presented in Figure \ref{fig:possible structure}. Future work will also explore generating input geometries based on desired performance, such by integrating the system in \cite{kyaw2025making}, or verifying structural performance, such as through built in FEA, as in \cite{smith_voxel_2025}. 

\subsection{Feeder Management}
As discussed in Section~\ref{sec:building-sequence}, in the current setup, each robot is paired with a unique Feed location, allowing independent operation. This constraint simplifies coordination, but limits scalability and task flexibility. Future work will explore decoupling the number of robots from the number of sources, enabling more dynamic task allocation. Related work  \cite{abdel-rahman_self-replicating_2022} has demonstrated that shared and flexible sourcing strategies can significantly improve resource utilization and system throughput.

Additionally, in the current system, we manually feed voxels to the robot. In a fully automated set up, this would be accomplished using e.g. conveyor belt to deliver voxels from their manufactured or stored location to the robots. 

\subsection{Robot Control}
As the system scales to multiple robots, hardware variability is expected to introduce differences in kinematics and precision. To ensure consistent behavior across platforms, we plan to develop a more adaptable control system with integrated per-robot calibration. This includes implementing a tunable inverse kinematics (IK) model, allowing each robot to compensate for mechanical deviations and maintain precise control during assembly.

Additionally, to fully realize voxel reconfiguration, the robots will need to eventually be able to disassemble voxel structures, which will require both additions to the robotic hardware as well as the path-planning tool. Eventually, we hope to be able to fully support the arbitrary disassembly and reconfiguration of a voxel structure using this system. 

\subsection{Voxel Material}

We use FFF printed voxel blocks in this work primarily because they are readily accessible and offer high performance along some loading directions. However, the anisotropies inherent in FFF plastic printing likely preclude its use for many practical applications, so future work will focus on developing alternate methods for pre-building the voxel blocks, such as through injection molded faces assembled together \cite{jenett_discretely_2020} or formed metallic facets \cite{parra2023kirigami}. Additionally, \cite{smith_voxel_2025}, \cite{cameron_discrete_2022}, and \cite{wang_voxelcopter_2023} present electrically active voxel systems, which could be integrated with this system in the future. Additionally, the system is likely mechanically anisotropic, and in future work we plan to develop additional tools simulated structure behavior, as well as more fully characterize the material system under bending and tensile loads. 

\section{Conclusion}

This work demonstrates a novel approach to scalable digital fabrication through the integration of architected lattice blocks, modular mobile robots, and a live digital twin system for coordinated assembly. By leveraging the strengths of small-scale digital fabrication to produce complex interlocking components, and combining them with simple, mobile robots capable of relative traversal and block placement, we enable the construction of meter-scale structures with reduced system complexity and cost. The hierarchical voxelization strategy, modular assembly robot design, and real-time simulation interface collectively contribute to a flexible and scalable assembly pipeline. Our validation of the system through the successful fabrication of meter-scale structures highlights its potential for broader application, such as for architecture, infrastructure, or in-space assembly and manufacturing. This work establishes groundwork for further exploring multi-robot collaboration in hardware, the interplay between structural robotic systems, and, eventually, hierarchical and recursive robotic assembly.

\begin{acks}
This work was supported by MIT Center for Bits and Atoms Consortia funding. 
\end{acks}

\bibliographystyle{ACM-Reference-Format}
\bibliography{SCF25}


\begin{thebibliography}{56}

%%% the Web address.

\ifx \showCODEN    \undefined \def \showCODEN     #1{\unskip}     \fi
\ifx \showDOI      \undefined \def \showDOI       #1{#1}\fi
\ifx \showISBNx    \undefined \def \showISBNx     #1{\unskip}     \fi
\ifx \showISBNxiii \undefined \def \showISBNxiii  #1{\unskip}     \fi
\ifx \showISSN     \undefined \def \showISSN      #1{\unskip}     \fi
\ifx \showLCCN     \undefined \def \showLCCN      #1{\unskip}     \fi
\ifx \shownote     \undefined \def \shownote      #1{#1}          \fi
\ifx \showarticletitle \undefined \def \showarticletitle #1{#1}   \fi
\ifx \showURL      \undefined \def \showURL       {\relax}        \fi
% invisible to TeX
\providecommand\bibfield[2]{#2}
\providecommand\bibinfo[2]{#2}
\providecommand\natexlab[1]{#1}
\providecommand\showeprint[2][]{arXiv:#2}

\bibitem[Abdel-Rahman et~al\mbox{.}(2022)]%
        {abdel-rahman_self-replicating_2022}
\bibfield{author}{\bibinfo{person}{Amira Abdel-Rahman}, \bibinfo{person}{Christopher Cameron}, \bibinfo{person}{Benjamin Jenett}, \bibinfo{person}{Miana Smith}, {and} \bibinfo{person}{Neil Gershenfeld}.} \bibinfo{year}{2022}\natexlab{}.
\newblock \showarticletitle{Self-replicating hierarchical modular robotic swarms}.
\newblock \bibinfo{journal}{\emph{Communications Engineering}} \bibinfo{volume}{1}, \bibinfo{number}{1} (\bibinfo{date}{November} \bibinfo{year}{2022}), \bibinfo{pages}{35}.
\newblock
\showISSN{2731-3395}
\urldef\tempurl%
\url{https://doi.org/10.1038/s44172-022-00034-3}
\showDOI{\tempurl}


\bibitem[Abdullah et~al\mbox{.}(2024)]%
        {10.1145/3639473.3665787}
\bibfield{author}{\bibinfo{person}{Muhammad Abdullah}, \bibinfo{person}{Laurenz Seidel}, \bibinfo{person}{Ben Wernicke}, \bibinfo{person}{Mehdi Gouasmi}, \bibinfo{person}{Anton Hackl}, \bibinfo{person}{Thomas Kern}, \bibinfo{person}{Conrad Lempert}, \bibinfo{person}{Clara Lempert}, \bibinfo{person}{David Bizer}, \bibinfo{person}{Wieland Storch}, \bibinfo{person}{Chiao Fang}, {and} \bibinfo{person}{Patrick Baudisch}.} \bibinfo{year}{2024}\natexlab{}.
\newblock \showarticletitle{PopCore: Personal Fabrication of 3D Foamcore Models for Professional High-Quality Applications in Design and Architecture}. In \bibinfo{booktitle}{\emph{Proceedings of the 9th ACM Symposium on Computational Fabrication}} (Aarhus, Denmark) \emph{(\bibinfo{series}{SCF '24})}. \bibinfo{publisher}{Association for Computing Machinery}, \bibinfo{address}{New York, NY, USA}, Article \bibinfo{articleno}{6}, \bibinfo{numpages}{14}~pages.
\newblock
\showISBNx{9798400704963}
\urldef\tempurl%
\url{https://doi.org/10.1145/3639473.3665787}
\showDOI{\tempurl}


\bibitem[Abdullah et~al\mbox{.}(2022)]%
        {10.1145/3526113.3545618}
\bibfield{author}{\bibinfo{person}{Muhammad Abdullah}, \bibinfo{person}{Romeo Sommerfeld}, \bibinfo{person}{Bjarne Sievers}, \bibinfo{person}{Leonard Geier}, \bibinfo{person}{Jonas Noack}, \bibinfo{person}{Marcus Ding}, \bibinfo{person}{Christoph Thieme}, \bibinfo{person}{Laurenz Seidel}, \bibinfo{person}{Lukas Fritzsche}, \bibinfo{person}{Erik Langenhan}, \bibinfo{person}{Oliver Adameck}, \bibinfo{person}{Moritz Dzingel}, \bibinfo{person}{Thomas Kern}, \bibinfo{person}{Martin Taraz}, \bibinfo{person}{Conrad Lempert}, \bibinfo{person}{Shohei Katakura}, \bibinfo{person}{Hany~Mohsen Elhassany}, \bibinfo{person}{Thijs Roumen}, {and} \bibinfo{person}{Patrick Baudisch}.} \bibinfo{year}{2022}\natexlab{}.
\newblock \showarticletitle{HingeCore: Laser-Cut Foamcore for Fast Assembly}. In \bibinfo{booktitle}{\emph{Proceedings of the 35th Annual ACM Symposium on User Interface Software and Technology}} (Bend, OR, USA) \emph{(\bibinfo{series}{UIST '22})}. \bibinfo{publisher}{Association for Computing Machinery}, \bibinfo{address}{New York, NY, USA}, Article \bibinfo{articleno}{10}, \bibinfo{numpages}{13}~pages.
\newblock
\showISBNx{9781450393201}
\urldef\tempurl%
\url{https://doi.org/10.1145/3526113.3545618}
\showDOI{\tempurl}


\bibitem[Agrawal et~al\mbox{.}(2015)]%
        {agrawal_protopiper_2015}
\bibfield{author}{\bibinfo{person}{Harshit Agrawal}, \bibinfo{person}{Udayan Umapathi}, \bibinfo{person}{Robert Kovacs}, \bibinfo{person}{Johannes Frohnhofen}, \bibinfo{person}{Hsiang-Ting Chen}, \bibinfo{person}{Stefanie Mueller}, {and} \bibinfo{person}{Patrick Baudisch}.} \bibinfo{year}{2015}\natexlab{}.
\newblock \showarticletitle{Protopiper: {Physically} {Sketching} {Room}-{Sized} {Objects} at {Actual} {Scale}}. In \bibinfo{booktitle}{\emph{Proceedings of the 28th {Annual} {ACM} {Symposium} on {User} {Interface} {Software} \& {Technology}}} \emph{(\bibinfo{series}{{UIST} '15})}. \bibinfo{publisher}{Association for Computing Machinery}, \bibinfo{address}{New York, NY, USA}, \bibinfo{pages}{427--436}.
\newblock
\showISBNx{978-1-4503-3779-3}
\urldef\tempurl%
\url{https://doi.org/10.1145/2807442.2807505}
\showDOI{\tempurl}


\bibitem[Apolinarska et~al\mbox{.}(2016)]%
        {apolinarska2016sequential}
\bibfield{author}{\bibinfo{person}{Aleksandra~Anna Apolinarska}, \bibinfo{person}{Michael Knauss}, \bibinfo{person}{Fabio Gramazio}, {and} \bibinfo{person}{Matthias Kohler}.} \bibinfo{year}{2016}\natexlab{}.
\newblock \showarticletitle{The Sequential Roof}.
\newblock In \bibinfo{booktitle}{\emph{Advancing Wood Architecture} (\bibinfo{edition}{1st edition} ed.)}. \bibinfo{publisher}{Routledge}, \bibinfo{pages}{14}.
\newblock
\showISBNx{9781315678825}
\newblock
\shownote{eBook}.


\bibitem[Apolinarska et~al\mbox{.}(2021)]%
        {APOLINARSKA2021103569}
\bibfield{author}{\bibinfo{person}{Aleksandra~Anna Apolinarska}, \bibinfo{person}{Matteo Pacher}, \bibinfo{person}{Hui Li}, \bibinfo{person}{Nicholas Cote}, \bibinfo{person}{Rafael Pastrana}, \bibinfo{person}{Fabio Gramazio}, {and} \bibinfo{person}{Matthias Kohler}.} \bibinfo{year}{2021}\natexlab{}.
\newblock \showarticletitle{Robotic assembly of timber joints using reinforcement learning}.
\newblock \bibinfo{journal}{\emph{Automation in Construction}}  \bibinfo{volume}{125} (\bibinfo{year}{2021}), \bibinfo{pages}{103569}.
\newblock
\showISSN{0926-5805}
\urldef\tempurl%
\url{https://doi.org/10.1016/j.autcon.2021.103569}
\showDOI{\tempurl}


\bibitem[Batikha et~al\mbox{.}(2022)]%
        {batikha_3d_2022}
\bibfield{author}{\bibinfo{person}{Mustafa Batikha}, \bibinfo{person}{Rahul Jotangia}, \bibinfo{person}{Mohamad~Yasser Baaj}, {and} \bibinfo{person}{Ibrahim Mousleh}.} \bibinfo{year}{2022}\natexlab{}.
\newblock \showarticletitle{{3D} concrete printing for sustainable and economical construction: {A} comparative study}.
\newblock \bibinfo{journal}{\emph{Automation in Construction}}  \bibinfo{volume}{134} (\bibinfo{date}{February} \bibinfo{year}{2022}), \bibinfo{pages}{104087}.
\newblock
\showISSN{0926-5805}
\urldef\tempurl%
\url{https://doi.org/10.1016/j.autcon.2021.104087}
\showDOI{\tempurl}


\bibitem[Baudisch et~al\mbox{.}(2019)]%
        {baudisch_kyub_2019}
\bibfield{author}{\bibinfo{person}{Patrick Baudisch}, \bibinfo{person}{Arthur Silber}, \bibinfo{person}{Yannis Kommana}, \bibinfo{person}{Milan Gruner}, \bibinfo{person}{Ludwig Wall}, \bibinfo{person}{Kevin Reuss}, \bibinfo{person}{Lukas Heilman}, \bibinfo{person}{Robert Kovacs}, \bibinfo{person}{Daniel Rechlitz}, {and} \bibinfo{person}{Thijs Roumen}.} \bibinfo{year}{2019}\natexlab{}.
\newblock \showarticletitle{Kyub: {A} {3D} {Editor} for {Modeling} {Sturdy} {Laser}-{Cut} {Objects}}. In \bibinfo{booktitle}{\emph{Proceedings of the 2019 {CHI} {Conference} on {Human} {Factors} in {Computing} {Systems}}}. \bibinfo{publisher}{ACM}, \bibinfo{address}{Glasgow Scotland Uk}, \bibinfo{pages}{1--12}.
\newblock
\showISBNx{978-1-4503-5970-2}
\urldef\tempurl%
\url{https://doi.org/10.1145/3290605.3300796}
\showDOI{\tempurl}


\bibitem[Bhundiya and Cordero(2023)]%
        {BHUNDIYA-bendforming}
\bibfield{author}{\bibinfo{person}{Harsh~G. Bhundiya} {and} \bibinfo{person}{Zachary~C. Cordero}.} \bibinfo{year}{2023}\natexlab{}.
\newblock \showarticletitle{Bend-Forming: A CNC deformation process for fabricating 3D wireframe structures}.
\newblock \bibinfo{journal}{\emph{Additive Manufacturing Letters}}  \bibinfo{volume}{6} (\bibinfo{year}{2023}), \bibinfo{pages}{100146}.
\newblock
\showISSN{2772-3690}
\urldef\tempurl%
\url{https://doi.org/10.1016/j.addlet.2023.100146}
\showDOI{\tempurl}


\bibitem[Bodea et~al\mbox{.}(2022)]%
        {bodea_additive_2022}
\bibfield{author}{\bibinfo{person}{Serban Bodea}, \bibinfo{person}{Pascal Mindermann}, \bibinfo{person}{Götz~T. Gresser}, {and} \bibinfo{person}{Achim Menges}.} \bibinfo{year}{2022}\natexlab{}.
\newblock \showarticletitle{Additive {Manufacturing} of {Large} {Coreless} {Filament} {Wound} {Composite} {Elements} for {Building} {Construction}}.
\newblock \bibinfo{journal}{\emph{3D Printing and Additive Manufacturing}} \bibinfo{volume}{9}, \bibinfo{number}{3} (\bibinfo{date}{June} \bibinfo{year}{2022}), \bibinfo{pages}{145--160}.
\newblock
\showISSN{2329-7662}
\urldef\tempurl%
\url{https://doi.org/10.1089/3dp.2020.0346}
\showDOI{\tempurl}
\newblock
\shownote{Publisher: Mary Ann Liebert, Inc., publishers}.


\bibitem[Cameron et~al\mbox{.}(2022)]%
        {cameron_discrete_2022}
\bibfield{author}{\bibinfo{person}{Christopher~G. Cameron}, \bibinfo{person}{Zach Fredin}, {and} \bibinfo{person}{Neil Gershenfeld}.} \bibinfo{year}{2022}\natexlab{}.
\newblock \showarticletitle{Discrete {Assembly} of {Unmanned} {Aerial} {Systems}}. In \bibinfo{booktitle}{\emph{2022 {International} {Conference} on {Unmanned} {Aircraft} {Systems} ({ICUAS})}}. \bibinfo{publisher}{IEEE}, \bibinfo{address}{Dubrovnik, Croatia}, \bibinfo{pages}{339--344}.
\newblock
\showISBNx{978-1-66540-593-5}
\urldef\tempurl%
\url{https://doi.org/10.1109/ICUAS54217.2022.9836082}
\showDOI{\tempurl}


\bibitem[Cheung and Gershenfeld(2013)]%
        {cheung_reversibly_2013}
\bibfield{author}{\bibinfo{person}{Kenneth~C. Cheung} {and} \bibinfo{person}{Neil Gershenfeld}.} \bibinfo{year}{2013}\natexlab{}.
\newblock \showarticletitle{Reversibly {Assembled} {Cellular} {Composite} {Materials}}.
\newblock \bibinfo{journal}{\emph{Science}} \bibinfo{volume}{341}, \bibinfo{number}{6151} (\bibinfo{date}{September} \bibinfo{year}{2013}), \bibinfo{pages}{1219--1221}.
\newblock
\showISSN{0036-8075, 1095-9203}
\urldef\tempurl%
\url{https://doi.org/10.1126/science.1240889}
\showDOI{\tempurl}


\bibitem[COBOD(2025)]%
        {cobod_eu_nodate}
\bibfield{author}{\bibinfo{person}{COBOD}.} \bibinfo{year}{2025}\natexlab{}.
\newblock \bibinfo{title}{{EU} {COBOD} {BOD2} {Tech} {Specs}}.
\newblock
\newblock
\urldef\tempurl%
\url{https://cobod.com/wp-content/uploads/2022/06/eu-cobod-bod2-techspecs.pdf}
\showURL{%
\tempurl}


\bibitem[Construction-Robotics(2025)]%
        {noauthor_sam_nodate}
\bibfield{author}{\bibinfo{person}{Construction-Robotics}.} \bibinfo{year}{2025}\natexlab{}.
\newblock \bibinfo{title}{{SAM}}.
\newblock
\newblock
\urldef\tempurl%
\url{https://www.construction-robotics.com/sam-2/}
\showURL{%
\tempurl}


\bibitem[Formlabs(2023)]%
        {formlabs2023howto}
\bibfield{author}{\bibinfo{person}{Formlabs}.} \bibinfo{year}{2023}\natexlab{}.
\newblock \bibinfo{booktitle}{\emph{How to Create Models Larger than Your 3D Printer's Build Volume}}.
\newblock
\urldef\tempurl%
\url{https://formlabs.com/blog/how-to-create-models-larger-than-your-3d-printers-build-volume/?srsltid=AfmBOorRcjw2tRh1lKB0ZLuxpdlQneRp-MpEOIe1SXpBOsrqNYOpiKLh}
\showURL{%
\tempurl}
\newblock
\shownote{Accessed: 2025-07-17}.


\bibitem[Gharbia et~al\mbox{.}(2020)]%
        {GHARBIA2020101584}
\bibfield{author}{\bibinfo{person}{Marwan Gharbia}, \bibinfo{person}{Alice Chang-Richards}, \bibinfo{person}{Yuqian Lu}, \bibinfo{person}{Ray~Y. Zhong}, {and} \bibinfo{person}{Heng Li}.} \bibinfo{year}{2020}\natexlab{}.
\newblock \showarticletitle{Robotic technologies for on-site building construction: A systematic review}.
\newblock \bibinfo{journal}{\emph{Journal of Building Engineering}}  \bibinfo{volume}{32} (\bibinfo{year}{2020}), \bibinfo{pages}{101584}.
\newblock
\showISSN{2352-7102}
\urldef\tempurl%
\url{https://doi.org/10.1016/j.jobe.2020.101584}
\showDOI{\tempurl}


\bibitem[Goessens et~al\mbox{.}(2018)]%
        {goessens_feasibility_2018}
\bibfield{author}{\bibinfo{person}{Sébastien Goessens}, \bibinfo{person}{Caitlin Mueller}, {and} \bibinfo{person}{Pierre Latteur}.} \bibinfo{year}{2018}\natexlab{}.
\newblock \showarticletitle{Feasibility study for drone-based masonry construction of real-scale structures}.
\newblock \bibinfo{journal}{\emph{Automation in Construction}}  \bibinfo{volume}{94} (\bibinfo{date}{October} \bibinfo{year}{2018}), \bibinfo{pages}{458--480}.
\newblock
\showISSN{0926-5805}
\urldef\tempurl%
\url{https://doi.org/10.1016/j.autcon.2018.06.015}
\showDOI{\tempurl}


\bibitem[Gregg et~al\mbox{.}(2024)]%
        {gregg_ultralight_2024}
\bibfield{author}{\bibinfo{person}{Christine~E. Gregg}, \bibinfo{person}{Damiana Catanoso}, \bibinfo{person}{Olivia Irene~B. Formoso}, \bibinfo{person}{Irina Kostitsyna}, \bibinfo{person}{Megan~E. Ochalek}, \bibinfo{person}{Taiwo~J. Olatunde}, \bibinfo{person}{In~Won Park}, \bibinfo{person}{Frank~M. Sebastianelli}, \bibinfo{person}{Elizabeth~M. Taylor}, \bibinfo{person}{Greenfield~T. Trinh}, {and} \bibinfo{person}{Kenneth~C. Cheung}.} \bibinfo{year}{2024}\natexlab{}.
\newblock \showarticletitle{Ultralight, strong, and self-reprogrammable mechanical metamaterials}.
\newblock \bibinfo{journal}{\emph{Science Robotics}} \bibinfo{volume}{9}, \bibinfo{number}{86} (\bibinfo{date}{January} \bibinfo{year}{2024}), \bibinfo{pages}{eadi2746}.
\newblock
\urldef\tempurl%
\url{https://doi.org/10.1126/scirobotics.adi2746}
\showDOI{\tempurl}
\newblock
\shownote{Publisher: American Association for the Advancement of Science}.


\bibitem[Gregg et~al\mbox{.}(2018)]%
        {gregg_ultra-light_2018}
\bibfield{author}{\bibinfo{person}{Christine~E. Gregg}, \bibinfo{person}{Joseph~H. Kim}, {and} \bibinfo{person}{Kenneth~C. Cheung}.} \bibinfo{year}{2018}\natexlab{}.
\newblock \showarticletitle{Ultra-{Light} and {Scalable} {Composite} {Lattice} {Materials}}.
\newblock \bibinfo{journal}{\emph{Advanced Engineering Materials}} \bibinfo{volume}{20}, \bibinfo{number}{9} (\bibinfo{year}{2018}), \bibinfo{pages}{1800213}.
\newblock
\showISSN{1527-2648}
\urldef\tempurl%
\url{https://doi.org/10.1002/adem.201800213}
\showDOI{\tempurl}
\newblock
\shownote{\_eprint: https://onlinelibrary.wiley.com/doi/pdf/10.1002/adem.201800213}.


\bibitem[HadrianX(2025)]%
        {noauthor_hadrian_nodate}
\bibfield{author}{\bibinfo{person}{HadrianX}.} \bibinfo{year}{2025}\natexlab{}.
\newblock \bibinfo{title}{Hadrian {X}® {\textbar} {Outdoor} {Construction} \& {Bricklaying} {Robot} from {FBR}}.
\newblock
\newblock
\urldef\tempurl%
\url{https://www.fbr.com.au/view/hadrian-x}
\showURL{%
\tempurl}


\bibitem[Hauser et~al\mbox{.}(2020)]%
        {HAUSER2020103467}
\bibfield{author}{\bibinfo{person}{S. Hauser}, \bibinfo{person}{M. Mutlu}, \bibinfo{person}{P.-A. Léziart}, \bibinfo{person}{H. Khodr}, \bibinfo{person}{A. Bernardino}, {and} \bibinfo{person}{A.J. Ijspeert}.} \bibinfo{year}{2020}\natexlab{}.
\newblock \showarticletitle{Roombots extended: Challenges in the next generation of self-reconfigurable modular robots and their application in adaptive and assistive furniture}.
\newblock \bibinfo{journal}{\emph{Robotics and Autonomous Systems}}  \bibinfo{volume}{127} (\bibinfo{year}{2020}), \bibinfo{pages}{103467}.
\newblock
\showISSN{0921-8890}
\urldef\tempurl%
\url{https://doi.org/10.1016/j.robot.2020.103467}
\showDOI{\tempurl}


\bibitem[Hsu et~al\mbox{.}(2016)]%
        {hsu_application_2016}
\bibfield{author}{\bibinfo{person}{Allen Hsu}, \bibinfo{person}{Annjoe Wong-Foy}, \bibinfo{person}{Brian McCoy}, \bibinfo{person}{Cregg Cowan}, \bibinfo{person}{John Marlow}, \bibinfo{person}{Bryan Chavez}, \bibinfo{person}{Takao Kobayashi}, \bibinfo{person}{Don Shockey}, {and} \bibinfo{person}{Ron Pelrine}.} \bibinfo{year}{2016}\natexlab{}.
\newblock \showarticletitle{Application of micro-robots for building carbon fiber trusses}. In \bibinfo{booktitle}{\emph{2016 {International} {Conference} on {Manipulation}, {Automation} and {Robotics} at {Small} {Scales} ({MARSS})}}. \bibinfo{pages}{1--6}.
\newblock
\urldef\tempurl%
\url{https://doi.org/10.1109/MARSS.2016.7561729}
\showDOI{\tempurl}


\bibitem[Jenett et~al\mbox{.}(2019)]%
        {jenett_materialrobot_2019}
\bibfield{author}{\bibinfo{person}{Benjamin Jenett}, \bibinfo{person}{Amira Abdel-Rahman}, \bibinfo{person}{Kenneth Cheung}, {and} \bibinfo{person}{Neil Gershenfeld}.} \bibinfo{year}{2019}\natexlab{}.
\newblock \showarticletitle{Material–{Robot} {System} for {Assembly} of {Discrete} {Cellular} {Structures}}.
\newblock \bibinfo{journal}{\emph{IEEE Robotics and Automation Letters}} \bibinfo{volume}{4}, \bibinfo{number}{4} (\bibinfo{date}{October} \bibinfo{year}{2019}), \bibinfo{pages}{4019--4026}.
\newblock
\showISSN{2377-3766, 2377-3774}
\urldef\tempurl%
\url{https://doi.org/10.1109/LRA.2019.2930486}
\showDOI{\tempurl}


\bibitem[Jenett et~al\mbox{.}(2017)]%
        {jenett_digital_2017}
\bibfield{author}{\bibinfo{person}{Benjamin Jenett}, \bibinfo{person}{Sam Calisch}, \bibinfo{person}{Daniel Cellucci}, \bibinfo{person}{Nick Cramer}, \bibinfo{person}{Neil Gershenfeld}, \bibinfo{person}{Sean Swei}, {and} \bibinfo{person}{Kenneth~C. Cheung}.} \bibinfo{year}{2017}\natexlab{}.
\newblock \showarticletitle{Digital {Morphing} {Wing}: {Active} {Wing} {Shaping} {Concept} {Using} {Composite} {Lattice}-{Based} {Cellular} {Structures}}.
\newblock \bibinfo{journal}{\emph{Soft Robotics}} \bibinfo{volume}{4}, \bibinfo{number}{1} (\bibinfo{date}{March} \bibinfo{year}{2017}), \bibinfo{pages}{33--48}.
\newblock
\showISSN{2169-5172}
\urldef\tempurl%
\url{https://doi.org/10.1089/soro.2016.0032}
\showDOI{\tempurl}
\newblock
\shownote{Publisher: Mary Ann Liebert, Inc., publishers}.


\bibitem[Jenett et~al\mbox{.}(2020)]%
        {jenett_discretely_2020}
\bibfield{author}{\bibinfo{person}{Benjamin Jenett}, \bibinfo{person}{Christopher Cameron}, \bibinfo{person}{Filippos Tourlomousis}, \bibinfo{person}{Alfonso~Parra Rubio}, \bibinfo{person}{Megan Ochalek}, {and} \bibinfo{person}{Neil Gershenfeld}.} \bibinfo{year}{2020}\natexlab{}.
\newblock \showarticletitle{Discretely assembled mechanical metamaterials}.
\newblock \bibinfo{journal}{\emph{Science Advances}} \bibinfo{volume}{6}, \bibinfo{number}{47} (\bibinfo{date}{November} \bibinfo{year}{2020}), \bibinfo{pages}{eabc9943}.
\newblock
\showISSN{2375-2548}
\urldef\tempurl%
\url{https://doi.org/10.1126/sciadv.abc9943}
\showDOI{\tempurl}


\bibitem[Jenett and Cheung(2017)]%
        {jenett_bill-e_2017}
\bibfield{author}{\bibinfo{person}{Ben Jenett} {and} \bibinfo{person}{Kenneth Cheung}.} \bibinfo{year}{2017}\natexlab{}.
\newblock \showarticletitle{{BILL}-{E}: {Robotic} {Platform} for {Locomotion} and {Manipulation} of {Lightweight} {Space} {Structures}}. In \bibinfo{booktitle}{\emph{25th {AIAA}/{AHS} {Adaptive} {Structures} {Conference}}}. \bibinfo{publisher}{American Institute of Aeronautics and Astronautics}, \bibinfo{address}{Grapevine, Texas}.
\newblock
\showISBNx{978-1-62410-446-6}
\urldef\tempurl%
\url{https://doi.org/10.2514/6.2017-1876}
\showDOI{\tempurl}


\bibitem[Jenett et~al\mbox{.}(2018)]%
        {jenett_building_2018}
\bibfield{author}{\bibinfo{person}{Benjamin Jenett}, \bibinfo{person}{Neil Gershenfeld}, {and} \bibinfo{person}{Paul Guerrier}.} \bibinfo{year}{2018}\natexlab{}.
\newblock \showarticletitle{Building {Block}-{Based} {Assembly} of {Scalable} {Metallic} {Lattices}}. In \bibinfo{booktitle}{\emph{Volume 4: {Processes}}}. \bibinfo{publisher}{American Society of Mechanical Engineers}, \bibinfo{address}{College Station, Texas, USA}, \bibinfo{pages}{V004T03A053}.
\newblock
\showISBNx{978-0-7918-5138-8}
\urldef\tempurl%
\url{https://doi.org/10.1115/MSEC2018-6442}
\showDOI{\tempurl}


\bibitem[Jenett(2020)]%
        {jenett_discrete_2020}
\bibfield{author}{\bibinfo{person}{Benjamin(Benjamin~Eric) Jenett}.} \bibinfo{year}{2020}\natexlab{}.
\newblock \emph{\bibinfo{title}{Discrete mechanical metamaterials}}.
\newblock Thesis. \bibinfo{school}{Massachusetts Institute of Technology}.
\newblock
\urldef\tempurl%
\url{https://dspace.mit.edu/handle/1721.1/130610}
\showURL{%
\tempurl}
\newblock
\shownote{Accepted: 2021-05-14T16:29:23Z}.


\bibitem[Kovacs et~al\mbox{.}(2017)]%
        {kovacs_trussfab_2017}
\bibfield{author}{\bibinfo{person}{Robert Kovacs}, \bibinfo{person}{Anna Seufert}, \bibinfo{person}{Ludwig Wall}, \bibinfo{person}{Hsiang-Ting Chen}, \bibinfo{person}{Florian Meinel}, \bibinfo{person}{Willi Müller}, \bibinfo{person}{Sijing You}, \bibinfo{person}{Maximilian Brehm}, \bibinfo{person}{Jonathan Striebel}, \bibinfo{person}{Yannis Kommana}, \bibinfo{person}{Alexander Popiak}, \bibinfo{person}{Thomas Bläsius}, {and} \bibinfo{person}{Patrick Baudisch}.} \bibinfo{year}{2017}\natexlab{}.
\newblock \showarticletitle{{TrussFab}: {Fabricating} {Sturdy} {Large}-{Scale} {Structures} on {Desktop} {3D} {Printers}}. In \bibinfo{booktitle}{\emph{Proceedings of the 2017 {CHI} {Conference} on {Human} {Factors} in {Computing} {Systems}}} \emph{(\bibinfo{series}{{CHI} '17})}. \bibinfo{publisher}{Association for Computing Machinery}, \bibinfo{address}{New York, NY, USA}, \bibinfo{pages}{2606--2616}.
\newblock
\showISBNx{978-1-4503-4655-9}
\urldef\tempurl%
\url{https://doi.org/10.1145/3025453.3026016}
\showDOI{\tempurl}


\bibitem[Kyaw et~al\mbox{.}(2025)]%
        {kyaw2025making}
\bibfield{author}{\bibinfo{person}{Alexander~Htet Kyaw}, \bibinfo{person}{Se~Hwan Jeon}, \bibinfo{person}{Miana Smith}, {and} \bibinfo{person}{Neil Gershenfeld}.} \bibinfo{year}{2025}\natexlab{}.
\newblock \bibinfo{title}{Making Physical Objects with Generative {AI} and Robotic Assembly: Considering Fabrication Constraints, Sustainability, Time, Functionality, and Accessibility}.
\newblock
\newblock
\showeprint[arxiv]{2504.19131}~[cs.RO]
\urldef\tempurl%
\url{https://arxiv.org/abs/2504.19131}
\showURL{%
\tempurl}
\newblock
\shownote{Preprint, presented at the CHI 2025 Workshop on Generative AI and Human--Computer Interaction}.


\bibitem[Leamon(2025)]%
        {Leamon2025}
\bibfield{author}{\bibinfo{person}{Sophie Leamon}.} \bibinfo{year}{2025}\natexlab{}.
\newblock \emph{\bibinfo{title}{Computational Design of Architected Lattices for Construction Applications}}.
\newblock Master's thesis. \bibinfo{school}{MASSACHUSETTS INSTITUTE OF TECHNOLOGY}, \bibinfo{address}{Cambridge, MA, USA}.
\newblock
\newblock
\shownote{Any additional info (optional)}.


\bibitem[Leder et~al\mbox{.}(2022)]%
        {leder_leveraging_2022}
\bibfield{author}{\bibinfo{person}{Samuel Leder}, \bibinfo{person}{HyunGyu Kim}, \bibinfo{person}{Ozgur~Salih Oguz}, \bibinfo{person}{Nicolas Kubail~Kalousdian}, \bibinfo{person}{Valentin~Noah Hartmann}, \bibinfo{person}{Achim Menges}, \bibinfo{person}{Marc Toussaint}, {and} \bibinfo{person}{Metin Sitti}.} \bibinfo{year}{2022}\natexlab{}.
\newblock \showarticletitle{Leveraging {Building} {Material} as {Part} of the {In}-{Plane} {Robotic} {Kinematic} {System} for {Collective} {Construction}}.
\newblock \bibinfo{journal}{\emph{Advanced Science}} \bibinfo{volume}{9}, \bibinfo{number}{24} (\bibinfo{year}{2022}), \bibinfo{pages}{2201524}.
\newblock
\showISSN{2198-3844}
\urldef\tempurl%
\url{https://doi.org/10.1002/advs.202201524}
\showDOI{\tempurl}
\newblock
\shownote{\_eprint: https://onlinelibrary.wiley.com/doi/pdf/10.1002/advs.202201524}.


\bibitem[Luban3D(2025)]%
        {luban3d2025}
\bibfield{author}{\bibinfo{person}{Luban3D}.} \bibinfo{year}{2025}\natexlab{}.
\newblock \bibinfo{booktitle}{\emph{Luban 3D}}.
\newblock
\urldef\tempurl%
\url{https://www.luban3d.com/}
\showURL{%
\tempurl}
\newblock
\shownote{Accessed: 2025-07-17}.


\bibitem[Ma et~al\mbox{.}(2020)]%
        {Ma-metal-frame}
\bibfield{author}{\bibinfo{person}{Z. Ma}, \bibinfo{person}{A. Walzer}, \bibinfo{person}{C. Schumacher}, \bibinfo{person}{R. Rust}, \bibinfo{person}{F. Gramazio}, \bibinfo{person}{M. Kohler}, {and} \bibinfo{person}{M. Bächer}.} \bibinfo{year}{2020}\natexlab{}.
\newblock \showarticletitle{Designing Robotically-Constructed Metal Frame Structures}.
\newblock \bibinfo{journal}{\emph{Computer Graphics Forum}} \bibinfo{volume}{39}, \bibinfo{number}{2} (\bibinfo{year}{2020}), \bibinfo{pages}{411--422}.
\newblock
\urldef\tempurl%
\url{https://doi.org/10.1111/cgf.13940}
\showDOI{\tempurl}
\showeprint{https://onlinelibrary.wiley.com/doi/pdf/10.1111/cgf.13940}


\bibitem[Marchment and Sanjayan(2020)]%
        {marchment_mesh_2020}
\bibfield{author}{\bibinfo{person}{Taylor Marchment} {and} \bibinfo{person}{Jay Sanjayan}.} \bibinfo{year}{2020}\natexlab{}.
\newblock \showarticletitle{Mesh reinforcing method for {3D} {Concrete} {Printing}}.
\newblock \bibinfo{journal}{\emph{Automation in Construction}}  \bibinfo{volume}{109} (\bibinfo{date}{January} \bibinfo{year}{2020}), \bibinfo{pages}{102992}.
\newblock
\showISSN{0926-5805}
\urldef\tempurl%
\url{https://doi.org/10.1016/j.autcon.2019.102992}
\showDOI{\tempurl}


\bibitem[Meisel et~al\mbox{.}(2022)]%
        {meisel_design_2022}
\bibfield{author}{\bibinfo{person}{Nicholas~A. Meisel}, \bibinfo{person}{Nathan Watson}, \bibinfo{person}{Sven~G. Bilén}, \bibinfo{person}{José~Pinto Duarte}, {and} \bibinfo{person}{Shadi Nazarian}.} \bibinfo{year}{2022}\natexlab{}.
\newblock \showarticletitle{Design and {System} {Considerations} for {Construction}-{Scale} {Concrete} {Additive} {Manufacturing} in {Remote} {Environments} via {Robotic} {Arm} {Deposition}}.
\newblock \bibinfo{journal}{\emph{3D Printing and Additive Manufacturing}} \bibinfo{volume}{9}, \bibinfo{number}{1} (\bibinfo{date}{February} \bibinfo{year}{2022}), \bibinfo{pages}{35--45}.
\newblock
\showISSN{2329-7662}
\urldef\tempurl%
\url{https://doi.org/10.1089/3dp.2020.0335}
\showDOI{\tempurl}
\newblock
\shownote{Publisher: Mary Ann Liebert, Inc., publishers}.


\bibitem[Melenbrink et~al\mbox{.}(2021)]%
        {melenbrink_autonomous_2021}
\bibfield{author}{\bibinfo{person}{Nathan Melenbrink}, \bibinfo{person}{Ariel Wang}, {and} \bibinfo{person}{Justin Werfel}.} \bibinfo{year}{2021}\natexlab{}.
\newblock \showarticletitle{An {Autonomous} {Vault}-{Building} {Robot} {System} for {Creating} {Spanning} {Structures}}. In \bibinfo{booktitle}{\emph{2021 {IEEE} {International} {Conference} on {Robotics} and {Automation} ({ICRA})}}. \bibinfo{pages}{7066--7072}.
\newblock
\urldef\tempurl%
\url{https://doi.org/10.1109/ICRA48506.2021.9561004}
\showDOI{\tempurl}
\newblock
\shownote{ISSN: 2577-087X}.


\bibitem[Menges et~al\mbox{.}(2017)]%
        {menges_fabricate_2017}
\bibfield{author}{\bibinfo{person}{Achim Menges}, \bibinfo{person}{Bob Sheil}, \bibinfo{person}{Ruairi Glynn}, {and} \bibinfo{person}{Marilena Skavara}.} \bibinfo{year}{2017}\natexlab{}.
\newblock \bibinfo{booktitle}{\emph{Fabricate 2017}}.
\newblock \bibinfo{publisher}{UCL Press}.
\newblock
\showISBNx{978-1-78735-001-4 978-1-78735-000-7}
\urldef\tempurl%
\url{https://doi.org/10.2307/j.ctt1n7qkg7}
\showDOI{\tempurl}


\bibitem[Neubert and Lipson(2016)]%
        {neubert_soldercubes_2016}
\bibfield{author}{\bibinfo{person}{Jonas Neubert} {and} \bibinfo{person}{Hod Lipson}.} \bibinfo{year}{2016}\natexlab{}.
\newblock \showarticletitle{Soldercubes: a self-soldering self-reconfiguring modular robot system}.
\newblock \bibinfo{journal}{\emph{Autonomous Robots}} \bibinfo{volume}{40}, \bibinfo{number}{1} (\bibinfo{date}{January} \bibinfo{year}{2016}), \bibinfo{pages}{139--158}.
\newblock
\showISSN{1573-7527}
\urldef\tempurl%
\url{https://doi.org/10.1007/s10514-015-9441-4}
\showDOI{\tempurl}


\bibitem[NXP(2021)]%
        {noauthor_i2c-bus_2021}
\bibfield{author}{\bibinfo{person}{NXP}.} \bibinfo{year}{2021}\natexlab{}.
\newblock \showarticletitle{{I2C}-bus specification and user manual}.
\newblock   \bibinfo{volume}{2021} (\bibinfo{year}{2021}).
\newblock


\bibitem[Park et~al\mbox{.}(2023)]%
        {park_soll-e_2023}
\bibfield{author}{\bibinfo{person}{In-Won Park}, \bibinfo{person}{Damiana Catanoso}, \bibinfo{person}{Olivia Formoso}, \bibinfo{person}{Christine Gregg}, \bibinfo{person}{Megan Ochalek}, \bibinfo{person}{Taiwo Olatunde}, \bibinfo{person}{Frank Sebastianelli}, \bibinfo{person}{Pascal Spino}, \bibinfo{person}{Elizabeth Taylor}, \bibinfo{person}{Greenfield Trinh}, {and} \bibinfo{person}{Kenneth Cheung}.} \bibinfo{year}{2023}\natexlab{}.
\newblock \showarticletitle{{SOLL}-{E}: {A} {Module} {Transport} and {Placement} {Robot} for {Autonomous} {Assembly} of {Discrete} {Lattice} {Structures}}. In \bibinfo{booktitle}{\emph{2023 {IEEE}/{RSJ} {International} {Conference} on {Intelligent} {Robots} and {Systems} ({IROS})}}. \bibinfo{publisher}{IEEE}, \bibinfo{address}{Detroit, MI, USA}, \bibinfo{pages}{10736--10741}.
\newblock
\showISBNx{978-1-66549-190-7}
\urldef\tempurl%
\url{https://doi.org/10.1109/IROS55552.2023.10341479}
\showDOI{\tempurl}


\bibitem[Parra~Rubio et~al\mbox{.}(2023)]%
        {parra_rubio_modular_2023}
\bibfield{author}{\bibinfo{person}{Alfonso Parra~Rubio}, \bibinfo{person}{Dixia Fan}, \bibinfo{person}{Benjamin Jenett}, \bibinfo{person}{José Del~Águila Ferrandis}, \bibinfo{person}{Filippos Tourlomousis}, \bibinfo{person}{Amira Abdel-Rahman}, \bibinfo{person}{David Preiss}, \bibinfo{person}{Jiri Zemánek}, \bibinfo{person}{Michael Triantafyllou}, {and} \bibinfo{person}{Neil Gershenfeld}.} \bibinfo{year}{2023}\natexlab{}.
\newblock \showarticletitle{Modular {Morphing} {Lattices} for {Large}-{Scale} {Underwater} {Continuum} {Robotic} {Structures}}.
\newblock \bibinfo{journal}{\emph{Soft Robotics}} \bibinfo{volume}{10}, \bibinfo{number}{4} (\bibinfo{date}{August} \bibinfo{year}{2023}), \bibinfo{pages}{724--736}.
\newblock
\showISSN{2169-5172, 2169-5180}
\urldef\tempurl%
\url{https://doi.org/10.1089/soro.2022.0117}
\showDOI{\tempurl}


\bibitem[PensaLabs(2025)]%
        {pensalabs2023}
\bibfield{author}{\bibinfo{person}{PensaLabs}.} \bibinfo{year}{2025}\natexlab{}.
\newblock \bibinfo{booktitle}{\emph{Pensa Labs}}.
\newblock
\urldef\tempurl%
\url{https://www.pensalabs.com/}
\showURL{%
\tempurl}
\newblock
\shownote{Accessed: 2025-07-17}.


\bibitem[Petersen et~al\mbox{.}(2011)]%
        {petersen2011termes}
\bibfield{author}{\bibinfo{person}{Kirstin Petersen}, \bibinfo{person}{Radhika Nagpal}, {and} \bibinfo{person}{Justin~K. Werfel}.} \bibinfo{year}{2011}\natexlab{}.
\newblock \showarticletitle{TERMES: An Autonomous Robotic System for Three-Dimensional Collective Construction}.
\newblock In \bibinfo{booktitle}{\emph{Robotics: Science and Systems VII}}, \bibfield{editor}{\bibinfo{person}{Hugh Durrant-Whyte}, \bibinfo{person}{Nicholas Roy}, {and} \bibinfo{person}{Pieter Abbeel}} (Eds.). \bibinfo{publisher}{MIT Press}, \bibinfo{address}{Cambridge, MA}.
\newblock
\urldef\tempurl%
\url{https://doi.org/10.15607/RSS.2011.VII.035}
\showDOI{\tempurl}


\bibitem[Petersen et~al\mbox{.}(2019)]%
        {petersen_review_2019}
\bibfield{author}{\bibinfo{person}{Kirstin~H. Petersen}, \bibinfo{person}{Nils Napp}, \bibinfo{person}{Robert Stuart-Smith}, \bibinfo{person}{Daniela Rus}, {and} \bibinfo{person}{Mirko Kovac}.} \bibinfo{year}{2019}\natexlab{}.
\newblock \showarticletitle{A review of collective robotic construction}.
\newblock \bibinfo{journal}{\emph{Science Robotics}} \bibinfo{volume}{4}, \bibinfo{number}{28} (\bibinfo{date}{March} \bibinfo{year}{2019}), \bibinfo{pages}{eaau8479}.
\newblock
\urldef\tempurl%
\url{https://doi.org/10.1126/scirobotics.aau8479}
\showDOI{\tempurl}
\newblock
\shownote{Publisher: American Association for the Advancement of Science}.


\bibitem[Pun et~al\mbox{.}(2025)]%
        {pun2025generatingphysicallystablebuildable}
\bibfield{author}{\bibinfo{person}{Ava Pun}, \bibinfo{person}{Kangle Deng}, \bibinfo{person}{Ruixuan Liu}, \bibinfo{person}{Deva Ramanan}, \bibinfo{person}{Changliu Liu}, {and} \bibinfo{person}{Jun-Yan Zhu}.} \bibinfo{year}{2025}\natexlab{}.
\newblock \bibinfo{title}{Generating Physically Stable and Buildable Brick Structures from Text}.
\newblock
\newblock
\showeprint[arxiv]{2505.05469}~[cs.CV]
\urldef\tempurl%
\url{https://arxiv.org/abs/2505.05469}
\showURL{%
\tempurl}


\bibitem[Roux et~al\mbox{.}(2023)]%
        {roux_life_2023}
\bibfield{author}{\bibinfo{person}{Charlotte Roux}, \bibinfo{person}{Kateryna Kuzmenko}, \bibinfo{person}{Nicolas Roussel}, \bibinfo{person}{Romain Mesnil}, {and} \bibinfo{person}{Adélaïde Feraille}.} \bibinfo{year}{2023}\natexlab{}.
\newblock \showarticletitle{Life cycle assessment of a concrete {3D} printing process}.
\newblock \bibinfo{journal}{\emph{The International Journal of Life Cycle Assessment}} \bibinfo{volume}{28}, \bibinfo{number}{1} (\bibinfo{date}{January} \bibinfo{year}{2023}), \bibinfo{pages}{1--15}.
\newblock
\showISSN{1614-7502}
\urldef\tempurl%
\url{https://doi.org/10.1007/s11367-022-02111-3}
\showDOI{\tempurl}


\bibitem[Rubio et~al\mbox{.}(2023)]%
        {parra2023kirigami}
\bibfield{author}{\bibinfo{person}{Alfonso~Parra Rubio}, \bibinfo{person}{Klara Mundilova}, \bibinfo{person}{David Preiss}, \bibinfo{person}{Erik Demaine}, {and} \bibinfo{person}{Neil Gershenfeld}.} \bibinfo{year}{2023}\natexlab{}.
\newblock \showarticletitle{Kirigami Corrugations: Strong, Modular and Programmable Plate Lattices}. In \bibinfo{booktitle}{\emph{Proceedings of the ASME 2023 International Design Engineering Technical Conferences and Computers and Information in Engineering Conference (IDETC-CIE2023)}}. \bibinfo{publisher}{ASME}.
\newblock
\urldef\tempurl%
\url{https://doi.org/10.1115/DETC2023-116481}
\showDOI{\tempurl}


\bibitem[Sass and Botha(2006)]%
        {sass2006instant}
\bibfield{author}{\bibinfo{person}{Lawrence Sass} {and} \bibinfo{person}{Marcel Botha}.} \bibinfo{year}{2006}\natexlab{}.
\newblock \showarticletitle{The Instant House: A Model of Design Production with Digital Fabrication}.
\newblock \bibinfo{journal}{\emph{International Journal of Architectural Computing}} \bibinfo{volume}{4}, \bibinfo{number}{4} (\bibinfo{year}{2006}), \bibinfo{pages}{109--123}.
\newblock
\urldef\tempurl%
\url{https://doi.org/10.1260/147807706779399015}
\showDOI{\tempurl}


\bibitem[Schaedler and Carter(2016)]%
        {schaedler_architected_2016}
\bibfield{author}{\bibinfo{person}{Tobias~A. Schaedler} {and} \bibinfo{person}{William~B. Carter}.} \bibinfo{year}{2016}\natexlab{}.
\newblock \showarticletitle{Architected {Cellular} {Materials}}.
\newblock \bibinfo{journal}{\emph{Annual Review of Materials Research}} \bibinfo{volume}{46}, \bibinfo{number}{1} (\bibinfo{year}{2016}), \bibinfo{pages}{187--210}.
\newblock
\urldef\tempurl%
\url{https://doi.org/10.1146/annurev-matsci-070115-031624}
\showDOI{\tempurl}
\newblock
\shownote{\_eprint: https://doi.org/10.1146/annurev-matsci-070115-031624}.


\bibitem[Skuric et~al\mbox{.}(2022)]%
        {simplefoc2022}
\bibfield{author}{\bibinfo{person}{Antun Skuric}, \bibinfo{person}{Hasan~Sinan Bank}, \bibinfo{person}{Richard Unger}, \bibinfo{person}{Owen Williams}, {and} \bibinfo{person}{David González-Reyes}.} \bibinfo{year}{2022}\natexlab{}.
\newblock \showarticletitle{SimpleFOC: A Field Oriented Control (FOC) Library for Controlling Brushless Direct Current (BLDC) and Stepper Motors}.
\newblock \bibinfo{journal}{\emph{Journal of Open Source Software}} \bibinfo{volume}{7}, \bibinfo{number}{74} (\bibinfo{year}{2022}), \bibinfo{pages}{4232}.
\newblock
\urldef\tempurl%
\url{https://doi.org/10.21105/joss.04232}
\showDOI{\tempurl}


\bibitem[Smith et~al\mbox{.}(2024)]%
        {smith_self-reconfigurable_2024}
\bibfield{author}{\bibinfo{person}{Miana Smith}, \bibinfo{person}{Amira Abdel-Rahman}, {and} \bibinfo{person}{Neil Gershenfeld}.} \bibinfo{year}{2024}\natexlab{}.
\newblock \showarticletitle{Self-{Reconfigurable} {Robots} for {Collaborative} {Discrete} {Lattice} {Assembly}}. In \bibinfo{booktitle}{\emph{2024 {IEEE} {International} {Conference} on {Robotics} and {Automation} ({ICRA})}}. \bibinfo{pages}{3624--3631}.
\newblock
\urldef\tempurl%
\url{https://doi.org/10.1109/ICRA57147.2024.10609866}
\showDOI{\tempurl}


\bibitem[Smith et~al\mbox{.}(2025)]%
        {smith_voxel_2025}
\bibfield{author}{\bibinfo{person}{Miana Smith}, \bibinfo{person}{Jack Forman}, \bibinfo{person}{Amira Abdel-Rahman}, \bibinfo{person}{Sophia Wang}, {and} \bibinfo{person}{Neil Gershenfeld}.} \bibinfo{year}{2025}\natexlab{}.
\newblock \showarticletitle{Voxel {Invention} {Kit}: {Reconfigurable} {Building} {Blocks} for {Prototyping} {Interactive} {Electronic} {Structures}}. In \bibinfo{booktitle}{\emph{Proceedings of the 2025 {CHI} {Conference} on {Human} {Factors} in {Computing} {Systems}}} \emph{(\bibinfo{series}{{CHI} '25})}. \bibinfo{publisher}{Association for Computing Machinery}, \bibinfo{address}{New York, NY, USA}, \bibinfo{pages}{1--15}.
\newblock
\showISBNx{9798400713941}
\urldef\tempurl%
\url{https://doi.org/10.1145/3706598.3713948}
\showDOI{\tempurl}


\bibitem[Smith(2023)]%
        {smith_recursive_2023}
\bibfield{author}{\bibinfo{person}{Miana~M. Smith}.} \bibinfo{year}{2023}\natexlab{}.
\newblock \emph{\bibinfo{title}{Recursive {Robotic} {Assemblers}}}.
\newblock Thesis. \bibinfo{school}{Massachusetts Institute of Technology}.
\newblock
\urldef\tempurl%
\url{https://dspace.mit.edu/handle/1721.1/152015}
\showURL{%
\tempurl}
\newblock
\shownote{Accepted: 2023-08-30T16:00:19Z}.


\bibitem[Wang et~al\mbox{.}(2023)]%
        {wang_voxelcopter_2023}
\bibfield{author}{\bibinfo{person}{Sophia Wang}, \bibinfo{person}{Miana Smith}, {and} \bibinfo{person}{Neil Gershenfeld}.} \bibinfo{year}{2023}\natexlab{}.
\newblock \showarticletitle{Voxelcopter: {Modular} {Autonomous} {Aerial} {Systems}}. In \bibinfo{booktitle}{\emph{Proceedings of the 8th {ACM} {Symposium} on {Computational} {Fabrication}}}. \bibinfo{publisher}{ACM}, \bibinfo{address}{New York City NY USA}, \bibinfo{pages}{1--2}.
\newblock
\showISBNx{9798400703195}
\urldef\tempurl%
\url{https://doi.org/10.1145/3623263.3629155}
\showDOI{\tempurl}


\bibitem[Yoon and Rus(2007)]%
        {yoon_shady3d_2007}
\bibfield{author}{\bibinfo{person}{Yeoreum Yoon} {and} \bibinfo{person}{Daniela Rus}.} \bibinfo{year}{2007}\natexlab{}.
\newblock \showarticletitle{{Shady3D}: {A} {Robot} that {Climbs} {3D} {Trusses}}. In \bibinfo{booktitle}{\emph{Proceedings 2007 {IEEE} {International} {Conference} on {Robotics} and {Automation}}}. \bibinfo{publisher}{IEEE}, \bibinfo{address}{Rome, Italy}, \bibinfo{pages}{4071--4076}.
\newblock
\showISBNx{978-1-4244-0602-9 978-1-4244-0601-2}
\urldef\tempurl%
\url{https://doi.org/10.1109/ROBOT.2007.364104}
\showDOI{\tempurl}
\newblock
\shownote{ISSN: 1050-4729}.


\end{thebibliography}

\end{document}